\documentclass[10pt,twocolumn,letterpaper]{article}

\usepackage{cvpr}
\usepackage{times}
\usepackage{epsfig}
\usepackage{graphicx}
\usepackage{amsmath}
\usepackage{amssymb}

\usepackage[pagebackref=true,breaklinks=true,letterpaper=true,colorlinks,bookmarks=false]{hyperref}

\usepackage[pass]{geometry}
\usepackage{multirow}
\usepackage{pdflscape}
\usepackage{rotating}
\usepackage{stfloats}
\newcommand{\mb}[1]{\mathbf{#1}}
\renewcommand{\th}{\text{th}}
\newcommand{\linetitle}[1]{\noindent\textbf{#1}~~}

\cvprfinalcopy 

\def\cvprPaperID{****} 

\begin{document}

\title{PA-GAN: Progressive Attention Generative Adversarial Network\\for Facial Attribute Editing}

\author{Zhenliang He$^{1,2}$, Meina Kan$^{1,2}$, Jichao Zhang$^1$, Shiguang Shan$^{1,2,3}$\\
$^1$ Key Laboratory of Intelligent Information Processing, ICT, CAS\\
$^2$ University of Chinese Academy of Sciences, Beijing, China\\
$^3$ Peng Cheng Laboratory, Shenzhen, China\\
{\tt\small \{zhenliang.he,jichao.zhang\}@vipl.ict.ac.cn, \{kanmeina,sgshan\}@ict.ac.cn}
\vspace{2mm}
}

\maketitle

\begin{abstract}
\vspace{-1mm}
Facial attribute editing aims to manipulate attributes on the human face, e.g., adding a mustache or changing the hair color.
Existing approaches suffer from a serious compromise between correct attribute generation and preservation of the other information such as identity and background, because they edit the attributes in the imprecise area.
To resolve this dilemma, we propose a progressive attention GAN (PA-GAN) for facial attribute editing.
In our approach, the editing is progressively conducted from high to low feature level while being constrained inside a proper attribute area by an attention mask at each level.
This manner prevents undesired modifications to the irrelevant regions from the beginning, and then the network can focus more on correctly generating the attributes within a proper boundary at each level.
As a result, our approach achieves correct attribute editing with irrelevant details much better preserved compared with the state-of-the-arts.
Codes are released at \url{https://github.com/LynnHo/PA-GAN-Tensorflow}.
\end{abstract}

\begin{figure}[!ht]
\begin{center}
\includegraphics[width=1\linewidth]{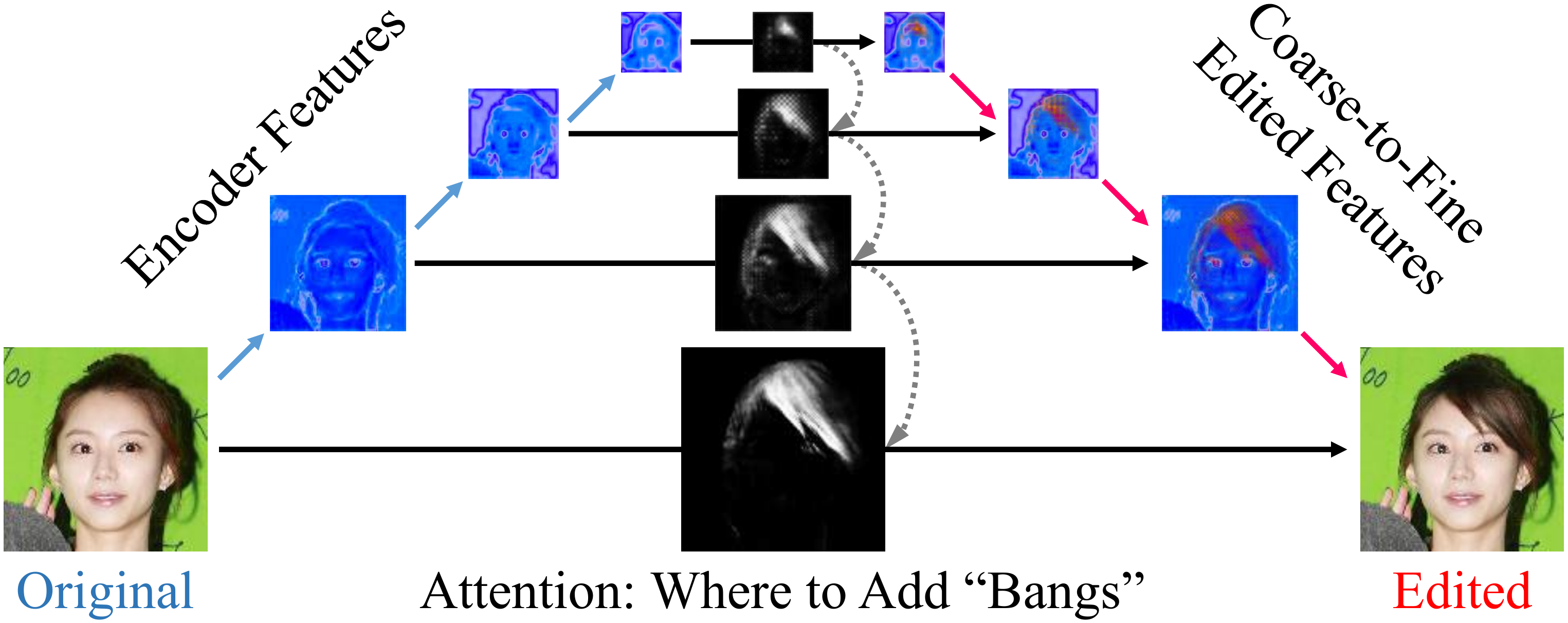}
\end{center}
\vspace{-1mm}
\caption{
A brief illustration of the proposed PA-GAN for facial attribute editing.
The editing is progressively conducted on the encoder feature maps inside the attention area from high to low feature level.
As the feature level gets lower (higher resolution), the attention mask gets more precise and the attribute editing becomes finer.
Edited parts of features on the right are shown in red.
}
\label{fig_first_view}
\end{figure}

\section{Introduction}

Facial attribute editing, i.e., manipulating attributes on the human face, has broad applications such as entertainment and data augmentation for other facial tasks like face recognition.
It is a sort of generative task and attracts more attention recently with the success of the generative models~\cite{goodfellow2014generative,kingma2013auto}.
A satisfactory facial attribute editing generally satisfies two criteria,
1) \textbf{attribute correctness}: the target attributes should correctly appear on the output image;
2) \textbf{irrelevance preservation}: the attribute-irrelevant parts, such as identity and background, should not be changed during editing.
However, these two objectives usually compromise with each other in existing methods.

\begin{figure*}[!t]
\begin{center}
\includegraphics[width=1\linewidth]{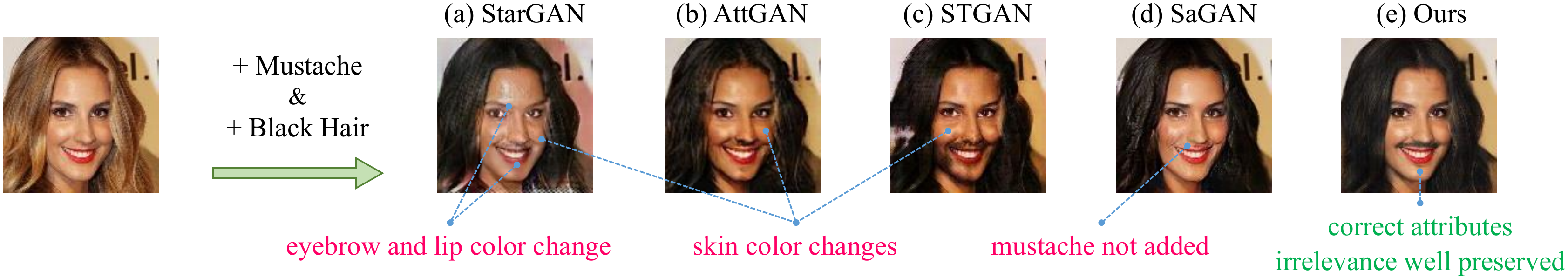}
\end{center}
\caption{
Illustration of the dilemma in existing methods and the advance of our method.
Our method edits the target attributes correctly with irrelevant parts well preserved, i.e., significantly alleviates the compromise between attribute correctness and irrelevance preservation.
}
\label{fig_first_comparison}
\end{figure*}

Recent state-of-the-art methods for facial attribute editing are mainly Generative Adversarial Network (GAN)~\cite{goodfellow2014generative} based ones, such as StarGAN~\cite{choi2018stargan}, AttGAN \cite{he2019attgan}, STGAN~\cite{liu2019stgan}, and SaGAN~\cite{zhang2018generative}.
Both StarGAN and AttGAN directly transform the input image to a whole new one with target attributes rather than edit the input image only within the proper attribute area, which inevitably results in undesired modifications of the irrelevant parts such as identity and background.
As illustrated by Fig.~\ref{fig_first_comparison}(a), StarGAN correctly edits the hair color and mustache but also changes the color of skin and lips.
In order to alleviate the compromise between attribute correctness and irrelevance preservation, STGAN~\cite{liu2019stgan} proposes selective transfer unit to adaptively select and modify the shortcut features instead of direct copy, achieving considerable improvement.
However, since STGAN also does not consider the editing area explicitly, it still cannot guarantee good preservation of the details in the attribute-irrelevant region.

In order to explicitly constrain the editing inside a reasonable area, SaGAN~\cite{zhang2018generative} applies global spatial attention on the input image to obtain a specific area and then conducts the attribute editing inside this area.
SaGAN successfully avoids most undesired modifications to the irrelevant region but at the cost of attribute correctness, because a straightforward attention mask only on the image cannot well handle multiple attributes with one model.
As illustrated by Fig.~\ref{fig_first_comparison}(d), SaGAN well preserves irrelevant parts but fails to add the desired ``Mustache''.

Overall, the most common problem of these methods is that they are unable to edit the attribute in the proper area, e.g., StarGAN edits more than expected while SaGAN edits less than expected, which results in the compromise between attribute correctness and irrelevance preservation.

To resolve this dilemma, in this work, we propose a progressive attention mechanism embedded in an encoder-decoder network (named as PA-GAN), which aims to precisely locate the editing area and meanwhile correctly edit the attributes inside.
As shown in Fig.~\ref{fig_first_view}, our model progressively conducts the attentive editing on the encoder feature maps from high to low feature level, regarding the input image as the lowest level feature.
At each level, there is an original (encoder) feature containing the original information such as background and identity, as well as a generated attribute feature containing the information of target attributes.
Then, our model learns an attention mask to blend the attribute feature into the original one, i.e., to conduct the attribute editing on the original feature inside a reasonable area indicated by the attention mask.

There are three main advantages of our approach:
1) conducting the attentive editing on the original (encoder) feature rather than generate a whole new feature like~\cite{choi2018stargan,he2019attgan,liu2019stgan,zhang2018generative}, can better preserve the original information of the irrelevant parts from various feature levels;
2) the network can focus more on correctly generating the attributes at each level given the proper attribute boundary;
3) the progressive manner separates the attentive editing into coarse-to-fine steps, which makes the task easier and produces better results.
Our contributions can be summarized as~follows:
\begin{itemize}
\item A novel generative model with progressive attention for facial attribute editing.
The proposed approach gradually conducts attentive attribute editing on the encoder features level by level, which forms a coarse-to-fine editing.

\item Attentive editing for multiple attributes with one model.
The overall attention mask is divided into multiple sub-masks, one for each desired attribute, which enables precise attention for multiple attributes with a single model.

\item Promising performance of facial attribute editing.
Comprehensive experiments on CelebA~\cite{liu2015deep} show that our approach can achieve high attribute correctness as well as satisfactory irrelevance preservation.
\end{itemize}

\section{Related Work}

\linetitle{Facial Attribute Editing}
Approaches for facial attribute editing can be categorized into two types: optimization-based approaches and learning-based approaches.
Optimization-based ones include CNAI~\cite{li2016convolutional} and DFI~\cite{upchurch2017deep}.
Both of these two approaches conduct the attribute editing by optimizing the image to narrow the gap between the deep feature of the image and a target attribute feature.
However, huge time consumption is their main drawback since the optimization process must be conducted for each test image.

Learning-based approaches are more popular~\cite{chen2018facelet,chen2019semantic,chen2019homomorphic,choi2018stargan,he2019attgan,kim2017unsupervised,lample2017fader,larsen2016autoencoding,li2016deep,liu2019stgan,perarnau2016invertible,romero2019smit,shen2017learning,sungatullina2018image,wu2019relgan,xiao2018dna,xiao2018elegant,yin2019instance,zhang2018generative,zhang2018sparsely,zhao2018modular,zhou2017genegan}, and approaches based on Generative Adversarial Networks (GANs)~\cite{goodfellow2014generative,mao2017least,arjovsky2017wasserstein,gulrajani2017improved} are the mainstream since GANs have achieved high fidelity image generation in various tasks~\cite{karras2018progressive,karras2019style,zhang2017stackgan,zhu2017unpaired}.
Earlier GAN based approaches include VAE/GAN~\cite{larsen2016autoencoding}, IcGAN~\cite{perarnau2016invertible}, and FaderNetwork~\cite{lample2017fader}.
VAE/GAN manipulates the attributes by moving the latent representation along the target attribute vector.
IcGAN and FaderNetwork extract an attribute-invariant representation of the input face and decode this representation specifying the target attributes.
The main problem of these methods is the constrained representation, which results in degraded performance, as illustrated in~\cite{he2019attgan}.
Latter, without any constraint on the latent, both StarGAN~\cite{choi2018stargan} and AttGAN~\cite{he2019attgan} improve the attribute editing performance by only employing the necessary adversarial loss, attribute classification loss, and the reconstruction loss.
STGAN~\cite{liu2019stgan} further improves the performance by employing an attribute difference indicator and selective transfer unit.
SaGAN~\cite{zhang2018generative} employs an attention manner that explicitly specifies an area on the image to edit one attribute with one model.
GeneGAN~\cite{zhou2017genegan}, DNAGAN~\cite{xiao2018dna}, ELEGANT~\cite{xiao2018elegant}, and Kim~\etal~\cite{kim2017unsupervised} swap attribute between two faces by exchanging the attribute-relevant latent codes, while GeoGAN~\cite{yin2019instance} achieves attribute swapping via a physical manner with the geometry-aware flow.
Other approaches without GAN also make considerable progress~\cite{chen2018facelet,chen2019semantic}. Facelet-Bank~\cite{chen2018facelet} employs a facelet module for each attribute to infer the feature deviation for attribute generation. Based on Facelet-Bank, Chen~\etal~\cite{chen2019semantic} further decomposes the feature deviation into different components with each corresponding to one kind of change, which achieves the fine-grained editing.

\linetitle{Generative Adversarial Networks (GANs)}
GANs~\cite{goodfellow2014generative,mao2017least,arjovsky2017wasserstein,gulrajani2017improved} make great progress in data generation recently.
Vanilla GAN~\cite{goodfellow2014generative} minimizes the Jensen-Shannon (JS) divergence between real and generated distribution by the adversarial training.
To stabilize the adversarial training, WGAN~\cite{arjovsky2017wasserstein}  optimizes the Wasserstein distance instead of JS divergence.
WGAN-GP~\cite{gulrajani2017improved} and WGAN-LP~\cite{petzka2018regularization} further improve WGAN by respectively employing gradient-penalty and Lipschitz penalty.
CGAN~\cite{mirza2014conditional} and AcGAN~\cite{odena2016conditional} are GAN extensions for conditional generation which can generate samples satisfying the given conditions.
In this work, WGAN-LP is used for stable training, and AcGAN is incorporated for attribute generation.

\section{Progressive Attention GAN (PA-GAN)}

\begin{figure*}[!ht]
\begin{center}
\includegraphics[width=0.9\linewidth]{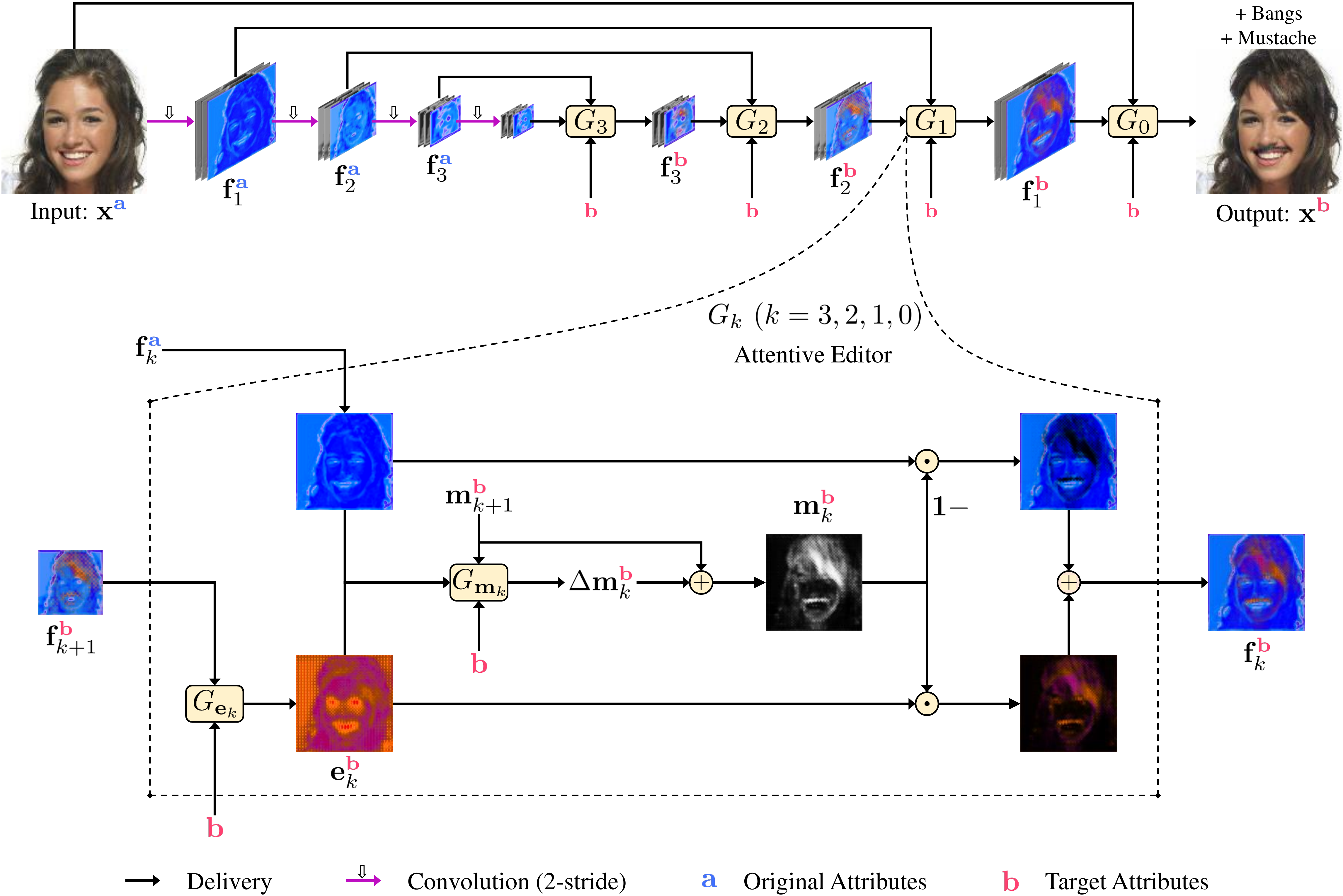}
\end{center}
\caption{
Overview of our PA-GAN.
PA-GAN progressively conducts attentive attribute editing on the original (encoder) features from high to low feature level, i.e., starting from $G_3$ to edit $\mb{f}^\mb{a}_3$ and ending at $G_0$ to edit $\mb{x}^\mb{a}$.
At each level, we design an Attentive Editor $G_k$ to edit the original feature $\mb{f}^\mb{a}_k$ with the attention mask $\mb{m}^\mb{b}_k$, producing the edited results $\mb{f}^\mb{b}_k$.
As the feature level gets lower, both the attention mask and the editing become more and more accurate.
The edited parts of $\mb{f}^\mb{b}_k$ are shown in red.
}
\label{fig_schema}
\end{figure*}

In this section, we introduce the details of the proposed Progressive Attention GAN.
Overall, as seen from the schema in Fig.~\ref{fig_schema}, our model progressively edits the original features (encoder features delivered by shortcut) in an attention manner from high to low feature level.
Furthermore, we extend this progressively attentive editing approach for multiple attributes with a single model.
Before introducing the details, we first clarify the notations for~convenience:
\begin{itemize}
\item For all $n$ attributes, $\mb{a}=[a_1, \cdots, a_i, \cdots, a_n]$ denotes the original attributes and $\mb{b}=[b_1, \cdots, b_i, \cdots, b_n]$ denotes the target attributes, where $a_i$ and $b_i$ are $1$/$0$ values denoting with/without the $i^\th$ attribute.
\item $\mb{x}^\mb{a}$ denotes an input image to be edited, and $\mb{x}^\mb{b}$ denotes the editing result with the target attributes $\mb{b}$, i.e., our goal can be formulated as $\mb{x}^\mb{a}\rightarrow \mb{x}^\mb{b}$.
\item
For the $k^\th$ level, $\mb{f}^\mb{a}_k$ denotes the original feature (encoder feature), and $\mb{f}^\mb{b}_k$ is the editing result at this level.
Then, the editing at the $k^\th$ level is formulated as the task: $\mb{f}^\mb{a}_k\rightarrow \mb{f}^\mb{b}_k$, i.e., to transform the ``original'' to the ``target''.
Specially, $\mb{f}^\mb{a}_0=\mb{x}^\mb{a}$ and $\mb{f}^\mb{b}_0=\mb{x}^\mb{b}$.
\end{itemize}

\subsection{Progressively Attentive Attribute Editing}\label{sec_progressive_attribute_editing}

Firstly, the input image $\mb{x}^\mb{a}$ is recursively encoded into features $\mb{f}^\mb{a}_k$ of different levels, formulated as follows,
\begin{align}
\mb{f}^\mb{a}_k & = E_k(\mb{f}^\mb{a}_{k-1}),~~k = 1, 2, 3, 4,\label{eq_fak}
\end{align}
where $\mb{f}^\mb{a}_0 = \mb{x}^\mb{a}$ and $E_k$ is the $k^\th$ encoder layer designed as a stride-2 convolution.
Via the encoder $E=[E_1, E_2, E_3, E_4]$, the original information of the input image, such as identity and background, are embedded in the encoder features $\{\mb{f}^\mb{a}_k\}_{k=0}^4$ of different aspects (different abstract levels).

Since the encoder features contain original information in various levels, we can directly conduct the editing on the encoder feature maps inside a proper attention area without affecting the irrelevant regions, which preserves the attribute-irrelevant details from various feature levels.
Moreover, given a proper editing area, the network can focus more on correctly generating the attribute at each level.
Furthermore, we can conduct the attentive editing progressively from high to low feature level, which forms a coarse-to-fine editing and produces better results.
Specifically, at each level, the original encoder feature $\mb{f}^\mb{a}_k$ is edited to be a new feature $\mb{f}^\mb{b}_k$, where the attribute is changed from $\mb{a}$ to $\mb{b}$, formulated as
\begin{align}
\mb{f}^\mb{b}_k & = G_k(\mb{f}^\mb{a}_k, \mb{f}^\mb{b}_{k+1}, \mb{b}-\mb{a}),~~k = 3, 2, 1, 0,\label{eq_Gk}
\end{align}
where $G_k$ is the Attentive Editor for the attentive attribute editing, which is introduced in detail in Sec.~\ref{sec_attention_editor};
$\mb{f}^\mb{b}_{k+1}$ is the result of the previous level, i.e., we conduct a finer editing borrowing the information of the previous coarse editing;
$\mb{b}-\mb{a}$ indicates the changing direction from attribute $\mb{a}$ to $\mb{b}$ following~\cite{liu2019stgan}.
Finally, when $k=0$ in Eq.~(\ref{eq_Gk}) with $\mb{x}^\mb{b} = \mb{f}^\mb{b}_{0}$ and $\mb{x}^\mb{a} = \mb{f}^\mb{a}_{0}$,
\begin{align}
\mb{x}^\mb{b} = G_{0}(\mb{x}^\mb{a}, \mb{f}^\mb{b}_{1}, \mb{b}-\mb{a}),
\end{align}
i.e., the attribute editing is finally finished on the input image $\mb{x}^\mb{a}$, obtaining a fine and accurate result $\mb{x}^\mb{b}$.

\subsection{Attentive Editor $\boldsymbol{G_k}$}\label{sec_attention_editor}

In this section, we introduce the Attentive Editor $G_{k}$ in Eq.~(\ref{eq_Gk}) in details.
Overall, $G_{k}$ conducts the attentive editing on the original feature $\mb{f}^\mb{a}_k$, aiming to restrict the attribute editing within a reasonable area at each level.
Specifically, the editing by $G_{k}$ is an alpha blending as~follows,
\begin{align}
\mb{f}^\mb{b}_k = \mb{m}^\mb{b}_k \cdot \mb{e}^\mb{b}_k + (\mb{1} - \mb{m}^\mb{b}_k) \cdot \mb{f}^\mb{a}_k ,\label{eq_blend}
\end{align}
where $\mb{m}^\mb{b}_k$ is the attention mask indicating a proper editing area, with value 1 for the attribute area and 0 for the irrelevant area;
$\mb{e}^\mb{b}_k$ is the generated attribute feature that contains the information of the target attribute.
Eq.~(\ref{eq_blend}) means that, instructed by $\mb{m}^\mb{b}_k$, the attribute part of the editing result $\mb{f}^\mb{b}_k$ comes from $\mb{e}^\mb{b}_k$ while the irrelevant part comes from the original $\mb{f}^\mb{a}_k$, i.e., the editing on $\mb{f}^\mb{a}_k$ is constrained in the area indicted by $\mb{m}^\mb{b}_k$.

More concretely, first, $\mb{e}^\mb{b}_k$ in Eq.~(\ref{eq_blend}) is an enhancement upon the editing result $\mb{f}^\mb{b}_{k+1}$of previous level, formulated as
\begin{align}
\mb{e}^\mb{b}_k = G_{\mb{e}_k}(\mb{f}^\mb{b}_{k+1}, \mb{b}-\mb{a}),\label{eq_ebk}
\end{align}
where $G_{\mb{e}_k}$ is a sub-network which enhances $\mb{f}^\mb{b}_{k+1}$ producing $\mb{e}^\mb{b}_k$ with better quality on target attribute.

Second, the attention mask $\mb{m}^\mb{b}_k$ in Eq.~(\ref{eq_blend}) is generated via a residual strategy, formulated as below,
\begin{align}
\mb{m}^\mb{b}_k &= \mb{m}^\mb{b}_{k+1} + \Delta \mb{m}^\mb{b}_k,~~k=3,2,1,0\label{eq_m}\\
\Delta \mb{m}^\mb{b}_k &= G_{\mb{m}_k}(\mb{f}^\mb{a}_k, \mb{e}^\mb{b}_k, \mb{m}^\mb{b}_{k+1}, \mb{b}-\mb{a}),\label{eq_dm}
\end{align}
where $\mb{m}^\mb{b}_4 = \mb{0}$, and $G_{\mb{m}_k}$ is in charge of predicting the residual to refine the mask $\mb{m}^\mb{b}_{k+1}$ of previous level, under the evidence of original feature $\mb{f}^\mb{a}_k$, the attribute feature $\mb{e}^\mb{b}_k$, and the previous mask $\mb{m}^\mb{b}_{k+1}$.
This residual strategy enables the attention mask to gradually become more and more precise from high to low feature level.
Specifically, starting from the highest feature level ($k=3$), the model captures the global information to obtain a coarse but robust attention mask.
As the feature level gets lower ($k=2,1,0$), the model absorbs more and more local information to refine the attention mask to be more and more precise.

Overall, $G_k$ produces the robust editing result $\mb{f}^\mb{b}_k$ guided by the attention mask $\mb{m}^\mb{b}_k$ as in Eq.~(\ref{eq_blend}).
In turn, at next level, $\mb{f}^\mb{b}_k$ helps to obtain finer attention mask $\mb{m}^\mb{b}_{k-1}$ by Eq. (\ref{eq_ebk})-(\ref{eq_dm}).
I.e., the editing and the attention learning iteratively benefit each other from high-level ($k=4$)  to low-level ($k=0$), therefore the overall attribute editing result becomes correct and precise progressively.

\subsection{Extension to Multiple Attributes}\label{sec_ext}

Although multiple attribute editing in a single model is achieved~\cite{choi2018stargan,he2019attgan,liu2019stgan}, it is not trivial to conduct \textit{attentively} multiple attribute editing.
Here, we extend the attention mask learning for multiple attributes by a divide-and-conquer strategy.
Specifically at each feature level, we learn $n$ distinct masks $[\mb{m}^{b_1}_k, \ldots, \mb{m}^{b_n}_k]$ for $n$ attributes, which is formulated as below,
\begin{align}
\mb{m}^{b_i}_k &= \mb{m}^{b_i}_{k+1} + \Delta \mb{m}^{b_i}_k,\label{eq_mi}\\
[\Delta\mb{m}^{b_1}_k, \ldots, \Delta\mb{m}^{b_n}_k] &= G_{\mb{m}_k}(\mb{f}^\mb{a}_k, \mb{e}^\mb{b}_k, \mb{m}^\mb{b}_{k+1}, \mb{b}-\mb{a}),\label{eq_dmi}
\end{align}
where $\mb{m}^{b_i}_k$ is the attention mask for the $i^\th$ attribute at the $k^\th$ level, and Eq.~(\ref{eq_mi})-(\ref{eq_dmi}) are adapted from Eq.~(\ref{eq_m})-(\ref{eq_dm}) for the multiple attribute setting.
Although there are $n$ masks for all $n$ attributes, we only need several of them which correspond to those attributes to be changed.
Specifically, if $b_i \neq a_i$, which means the $i^\th$ attribute needs to be changed from $a_i$ to $b_i$, then the corresponding mask $\mb{m}^{b_i}_k$ should be chosen; otherwise if $b_i = a_i$, which means the $i^\th$ attribute does not change, then the corresponding mask should be neglected.
Therefore, we choose the necessary masks and merged them into one mask, formulated as
\begin{align}
\mb{m}^\mb{b}_k = &\sum_i^n \delta(b_i \neq a_i) \cdot \mb{m}^{b_i}_k.
\end{align}
As can be seen, the learning of the attention mask for multiple attributes is divided into sub-tasks, with each learning a specific attention mask for a specific attribute, which reduces the difficulty as well as enhances the robustness of the mask prediction.
Here, only the way of obtaining the attention mask is different from that in Sec.~\ref{sec_attention_editor}, but the editing keeps the same as that in Eq.~(\ref{eq_blend})-(\ref{eq_ebk}).

\subsection{Objectives}

In this section, we introduce the objectives for training our model.
We employ the attribute prediction loss for correct attribute generation, as well as the adversarial loss for the generation fidelity.
Further, we design two mask losses for learning a more precise attention mask.

\linetitle{Attribute Prediction Loss}
If the attribute editing is correct, the edited image should be predicted to own the target attributes by an attribute predictor.
Therefore, we employ the attribute prediction loss to guide the correct generation of the target attributes, formulated as below,
\begin{align}
\min_{\{E_k\}, \{G_k\}} l_{att} =  \sum_i^n -[ & b_i \cdot \log(C_i(\mb{x}^\mb{b})) + \nonumber\\
                                               & (1-b_i) \cdot \log(1-C_i(\mb{x}^\mb{b}))],
\end{align}
where $C_i$ is an attribute classifier that predicts the probability of the $i^\th$ attribute to appear, and this loss is the summation of the binary cross entropy of all attributes.
Therefore, $E_k$ and $G_k$ will be penalized if the target attributes are not correctly generated on the output image $\mb{x}^\mb{b}$.
The attribute classifier $C_i$ is learnt with the real data as below,
\begin{align}
\min_{\{C_i\}} l_C =  \sum_i^n -[ & a_i \cdot \log(C_i(\mb{x}^\mb{a})) + \nonumber\\
                                  & (1-a_i) \cdot \log(1-C_i(\mb{x}^\mb{a}))],
\end{align}

\linetitle{Adversarial Loss}
We employ the adversarial loss for the generation fidelity, formulated as below,
\begin{align}
\max_{||D||_L\leq1} l^d_{adv}     & = \mathbb{E}[D(\mb{x}^\mb{a})] - \mathbb{E}[D(\mb{x}^\mb{b})],\label{eq_d}\\
\min_{\{E_k\}, \{G_k\}} l^g_{adv} & = -\mathbb{E}[D(\mb{x}^\mb{a})] + \mathbb{E}[D(\mb{x}^\mb{b})],\label{eq_g}
\end{align}
where $D$ is the discriminator constrained by 1-Lipschitz continuity following~\cite{petzka2018regularization}.
Eq.~(\ref{eq_d}) estimates the Wasserstein distance between the generated distribution and the real distribution, while Eq.~(\ref{eq_g}) minimizes this distance.
These two objectives are optimized iteratively and the generated distribution is optimally identical to the real one, i.e., the generated $\mb{x}^\mb{b}$ will look like a real image.

\linetitle{Attention Mask Losses}
We also constrain the sparsity of the attention masks in order to make them focus the limited value on the proper attribute region rather than the whole image, with the sparsity loss formulated as follows,
\begin{align}
\min_{\{E_k\}, \{G_k\}} l_{spa} &= \sum_k|\mb{m}^\mb{b}_k|_1\label{eq_lspa}
\end{align}
Besides, there exists a prior that some attributes have disjoint regions, e.g., attention masks of ``Bangs'' and ``Mustache'' definitely should not have overlap.
Therefore, we design an overlap loss to make the attention masks satisfy such prior.
Specifically, the element-wise multiplication of two masks for two disjoint attributes should be zero, otherwise there should be a penalty as follows,
\begin{align}
\min_{\{E_k\}, \{G_k\}} l_{ovl} &= \sum_{(i,j) \in S}|\mb{m}^{b_i}_0 \cdot \mb{m}^{b_j}_{0}|_1,\label{eq_lovl}
\end{align}
where $S$ is a predefined set containing pairs of attributes that have disjoint regions such as (``Blond Hair'', ``Mustache'') and (``Eyeglasses'', ``Beard''), and the full definition of $S$ can be found in the supplementary material.

\linetitle{Overall Objective}
Finally, we have an overall objective for the attribute editing network (including $E_k$ and $G_k$) as
\begin{align}
\min_{\{E_k\}, \{G_k\}} \lambda_1 l_{att} + \lambda_2 l^g_{adv} + \lambda_3 l_{spa} + \lambda_4 l_{ovl}.\label{eq_E_G}
\end{align}
Besides, the objective for the attribute classifier $C_i$ and the discriminator $D$ is
\begin{align}
\min_{\{C_i\}, ||D||_L\leq1} \lambda_5 l_C - \lambda_6 l^d_{adv},\label{eq_C_D}
\end{align}
where $C_i$ and $D$ share most layers except for the last two layers. $\lambda_1$, $\lambda_2$, $\lambda_3$, $\lambda_4$, $\lambda_5$, and $\lambda_6$ are the hyperparameters to balance the losses.

Note that there is no reconstruction loss for irrelevance preservation like previous methods~\cite{choi2018stargan,he2019attgan,liu2019stgan}, because the mask learning in our approach with the above objectives is precise enough to avoid the affection on the irrelevance.

\subsection{Differences from Related Methods}

Two closely related works are AttGAN~\cite{he2019attgan} and STGAN~\cite{liu2019stgan}, which also adapt encoder-decoder architecture for facial attribute editing.
Following the U-Net~\cite{ronneberger2015u} architecture, AttGAN directly appends the encoder features as a supplement to the decoder features for subsequent convolutions, while STGAN selectively transforms the encoder features before the concatenation.
Different from these concatenations, our approach progressively blends the attribute features into the encoder features guided by the attention mask.
Besides, note that our approach is different from the approaches for other tasks~\cite{seo2016progressive,kim2019progressive,zhang2018progressive} also called ``progressive attention''.
The ``attention'' in~\cite{seo2016progressive,kim2019progressive,zhang2018progressive} is to learn weights to re-weight the features for subsequent convolutions, and their ``progressive'' means to re-weight the features at each layer.
Differently, our ``attention'' is to locate the attribute region for blending, and our ``progressive'' means to learn coarse-to-fine attention with the residual learning and edit the attributes progressively.

\section{Experiments}

\linetitle{Dataset}
We adopt CelebA~\cite{liu2015deep} to evaluate the proposed PA-GAN.
CelebA contains 202,599 images with annotations of 40 binary attributes.
Following~\cite{he2019attgan,liu2019stgan}, we select thirteen attributes in all our experiments, including \textit{Bald, Bangs, Black Hair, Blond Hair, Brown Hair, Bushy Eyebrows, Eyeglasses, Gender, Mouth Open, Mustache, Beard, Pale Skin, and Age}.
Besides, 182637 images are used as the training set, and 19962 images are used as the testing set.

\linetitle{Competitors}
Recent state-of-the-art methods including StarGAN~\cite{choi2018stargan}, AttGAN \cite{he2019attgan}, STGAN~\cite{liu2019stgan} and SaGAN~\cite{zhang2018generative} are chosen as the competitors.
All these methods are trained and evaluated under the same protocol, using multi-attribute models.
Especially, we extend SaGAN to multiple attribute model since the original SaGAN can only handle one attribute with one model.

\linetitle{Implementation Details}
Loss weights in Eq.~(\ref{eq_E_G})-(\ref{eq_C_D}) are set as $\lambda_1=20$, $\lambda_2=1$, $\lambda_3=0.05$, $\lambda_4=1$, $\lambda_5=1$, and $\lambda_6=1$, which balances the magnitude of these losses to the same order.
All networks are optimized by Adam solver~\cite{kingma2015adam} ($\beta_1=0.5$, $\beta_2=0.999$).
All experiments are conducted on 128$\times$128 images following the default size of \cite{choi2018stargan,liu2019stgan}.
Please refer to the supplementary material for \textit{the network architectures}, and \textit{more higher resolution results}.

\subsection{Qualitative Analysis}

\begin{figure*}[!ht]
\begin{center}
\includegraphics[width=1\linewidth]{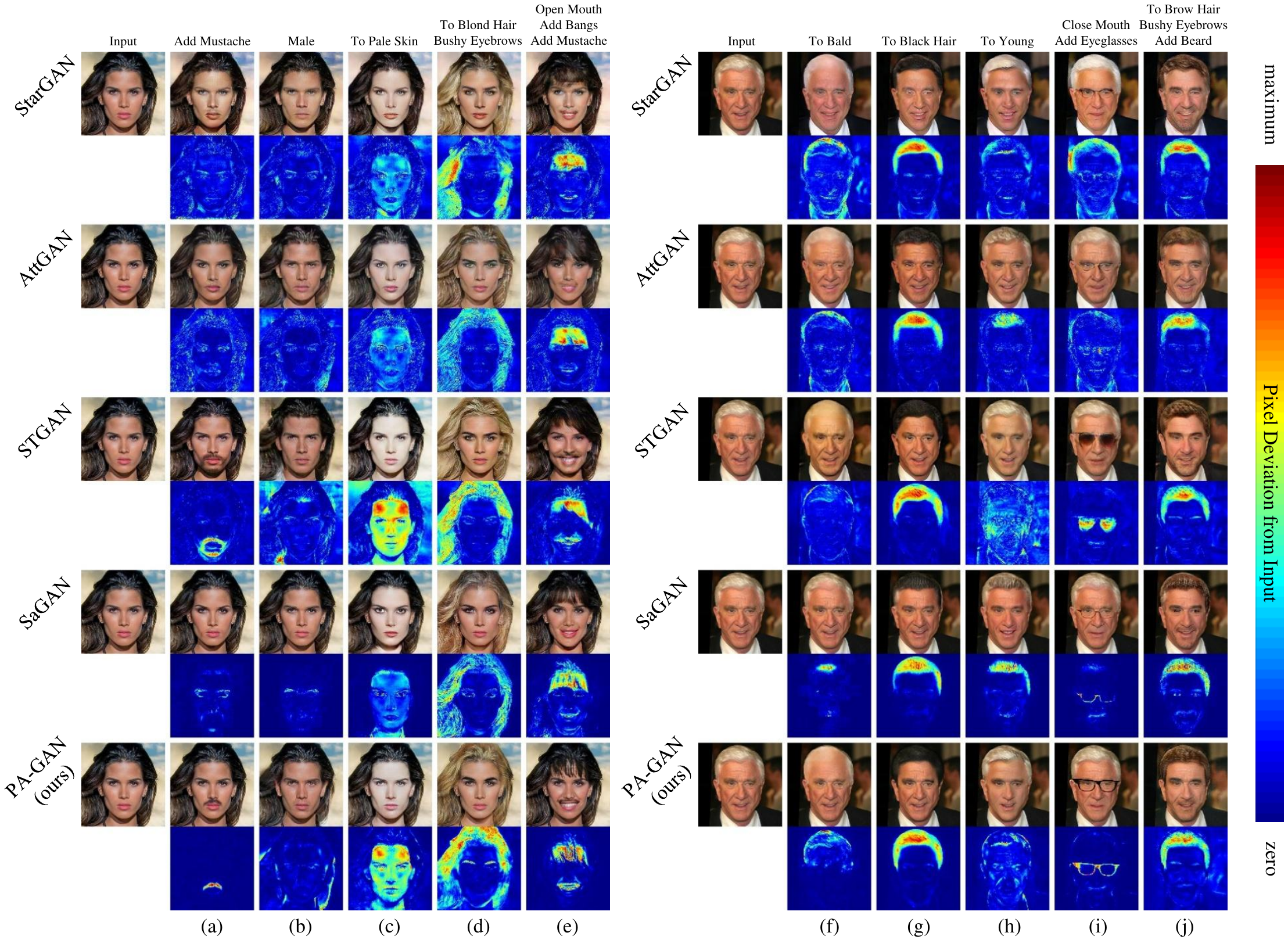}
\end{center}
\vspace{-3mm}
\caption{Visual comparisons among StarGAN~\cite{choi2018stargan}, AttGAN~\cite{he2019attgan}, STGAN~\cite{liu2019stgan}, SaGAN~\cite{zhang2018generative}, and the proposed PA-GAN. Colormaps below the editing results show the pixel deviation of the edited image from the input image.}
\label{fig_comparison}
\vspace{-4mm}
\end{figure*}

\linetitle{Visual Results}
Fig.~\ref{fig_comparison} shows the visual results of the competing methods, as well as the deviation colormaps showing where and how much the pixels differ from the input image.
As can be seen, StarGAN correctly edits the attributes; however, there are obvious undesired changes, e.g., the skin color changes in all situations and the shape of mouth changes when editing the \textit{Young} attribute (Fig.~\ref{fig_comparison}(h), row 1).
AttGAN fails to add \textit{Eyeglasses} (Fig.~\ref{fig_comparison}(i), row 3) and changes the skin color when adding \textit{Black Hair} (Fig.~\ref{fig_comparison}(g), row 3).
STGAN correctly edits the attributes; however, it also changes undesired parts, e.g., the background becomes white when changing the input to \textit{Pale Skin} (Fig.~\ref{fig_comparison}(c), row 5) and the skin color changes when editing \textit{Bald} (Fig.~\ref{fig_comparison}(f), row 5).
StarGAN, AttGAN, and STGAN produce undesired modifications to the irrelevant region, since they transform the input image to a whole new one without explicitly considering to edit the attributes in a proper area.
SaGAN has much cleaner colormaps than StarGAN, AttGAN, and STGAN; however, it fails on adding \textit{Mustache} (Fig.~\ref{fig_comparison}(a), row 7) and changing to \textit{Male} (Fig.~\ref{fig_comparison}(b), row 7), because it cannot handle multiple attributes in one model with only one global attention on the image.
As can be seen, there is obvious compromise between attribute correctness and irrelevance preservation in these methods.
In comparison, our PA-GAN correctly and naturally edits the attributes while the other details such as skin color, shape, and background are well preserved, demonstrating its effectiveness.

\begin{figure}[!ht]
\begin{center}
\includegraphics[width=1\linewidth]{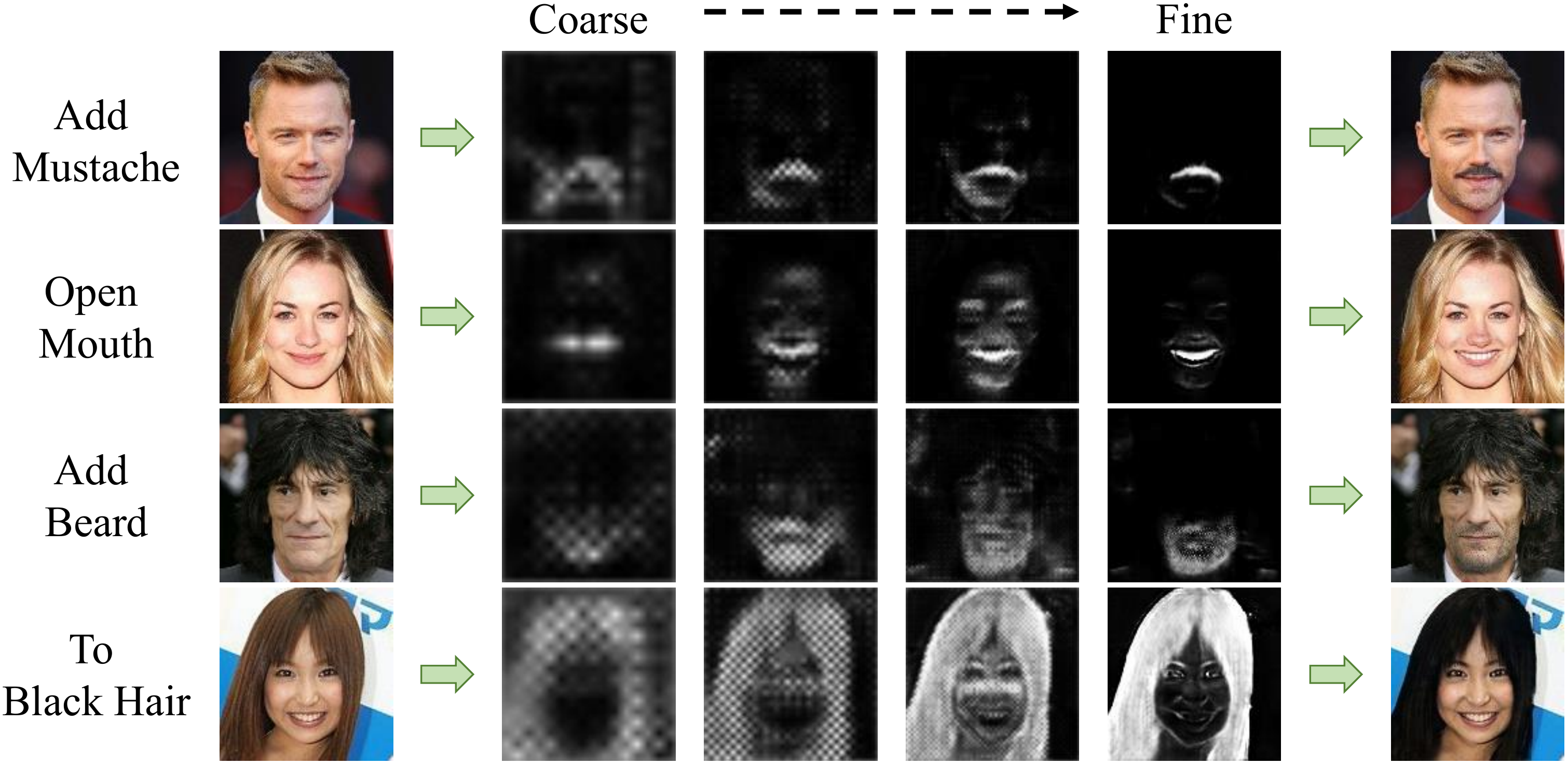}
\end{center}
\vspace{-3mm}
\caption{Attention mask from high to low feature level (left to right). We rescale the mask to the same size for better view.}
\label{fig_mask}
\vspace{-4mm}
\end{figure}

\begin{figure}[!ht]
\begin{center}
\includegraphics[width=1\linewidth]{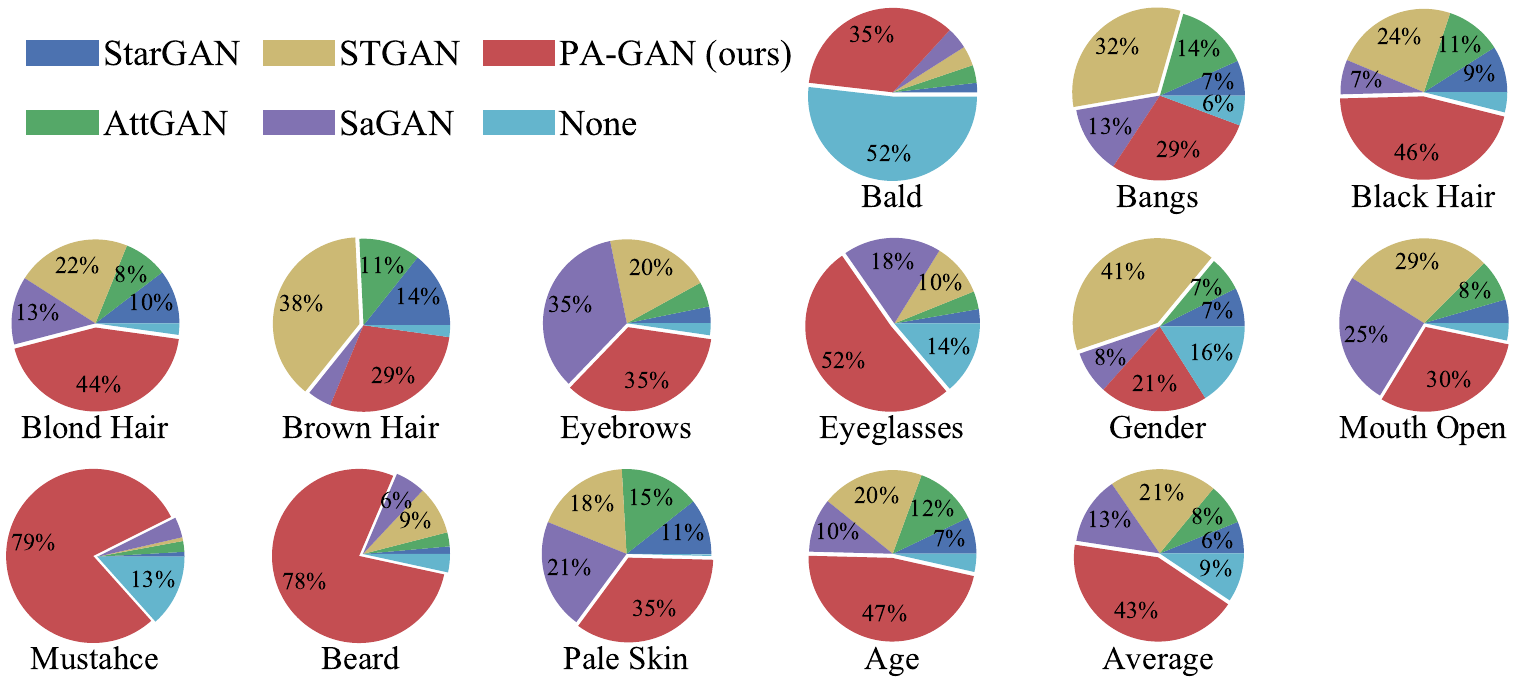}
\end{center}
\vspace{-3mm}
\caption{User study results which show the proportion of each method to be chosen as the best for each attribute. Values smaller than 5\% are omitted for clear display.}
\label{fig_user_study}
\vspace{-5mm}
\end{figure}

\linetitle{Coarse-to-Fine Attention}
Fig.~\ref{fig_mask} shows the attention masks at all levels.
As can be seen, the mask gradually becomes finer and more precise, which demonstrates the coarse-to-fine effect of our progressive attention approach.

\subsection{Quantitative Analysis}

\linetitle{User Study}
We conduct a user study to evaluate the proposed PA-GAN under human perception.
Specifically, we randomly choose 100-120 images for each attribute, and these images are the same for all methods.
Then for each image, 10 volunteers are asked to select the best editing result from all competing methods, according to 1) whether the attribute is correctly generated, and 2) whether the irrelevant parts are well preserved.
Specially, we also provide a choice of ``none of these methods performs well'' to investigate the hard attributes which cannot be well solved by all methods including ours.
Fig.~\ref{fig_user_study} shows the proportion of each method to be chosen as the best for each attribute averaged over all 10 volunteers.
As can be seen, the proposed PA-GAN is chosen as best for most attributes, demonstrating the superiority of PA-GAN to the competitors.
Besides, ``none of these methods performs well'' is chosen most for \textit{Bald} attribute.
One possible reason is the imbalance data distribution of \textit{Bald}: only 2\% of the data have \textit{Bald} attribute, while almost all of these samples are male.

\begin{figure}[!ht]
\begin{center}
\vspace{-3mm}
\includegraphics[width=1\linewidth]{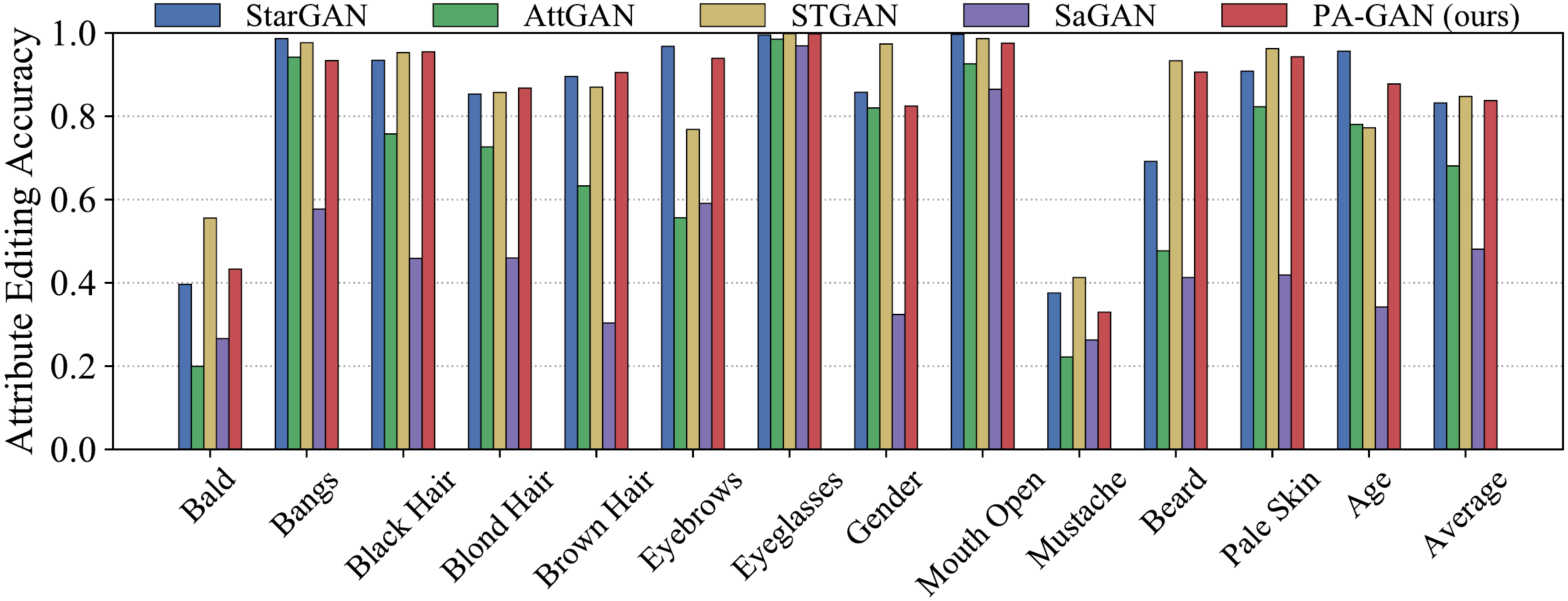}
\vspace{-5mm}
\caption{Attribute editing accuracy, higher is better.}
\label{fig_attribute_accuracy}
\end{center}
\end{figure}

\linetitle{Attribute Editing Accuracy}
Attribute editing accuracy is to evaluate whether a specified attribute correctly appears on the image.
We used a well trained attribute predictor ($95\%$ attribute prediction accuracy) to judge whether the attribute editing is correct.
Fig.~\ref{fig_attribute_accuracy} shows the attribute editing accuracy of all competing methods.
StarGAN, STGAN, and the proposed PA-GAN achieve comparable attribute editing accuracies, which are superior to AttGAN and SaGAN, demonstrating that our PA-GAN model can generate attributes with high correctness.
Although StarGAN and STGAN achieve high attribute editing accuracy, both of them affect the attribute irrelevant region while the proposed PA-GAN achieves much better irrelevance preservation, as illustrated in the visual results in Fig.~\ref{fig_comparison} and the irrelevance preservation error in the next paragraph.

\begin{figure}[!ht]
\begin{center}
\includegraphics[width=1\linewidth]{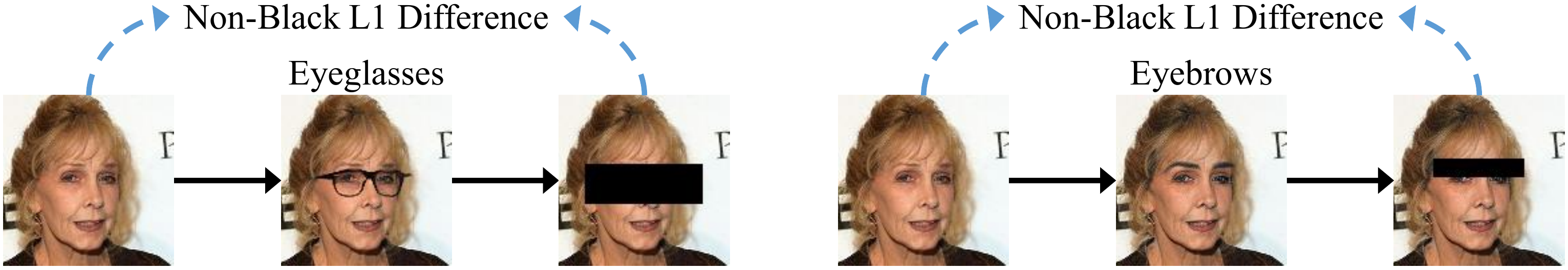}
\end{center}
\vspace{-2mm}
\caption{The irrelevance preservation error is calculated as the L1 difference of the non-black region between the editing result and the input image. The non-black region indicates those irrelevant regions that should be kept unchanged, which is predefined for each attribute since all faces are aligned.}
\label{fig_mask_for_appearance_change}
\end{figure}

\begin{figure}[!ht]
\begin{center}
\vspace{-5mm}
\includegraphics[width=1\linewidth]{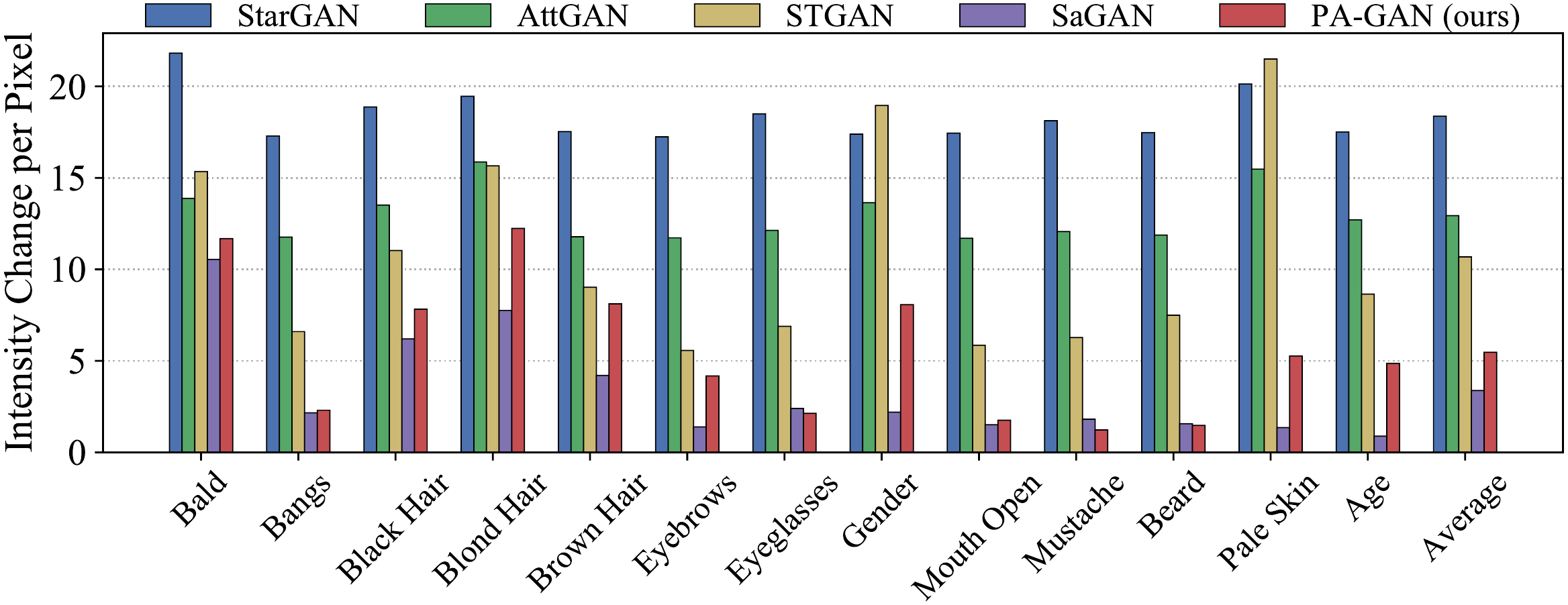}
\vspace{-5mm}
\caption{Irrelevance preservation error, lower is better.}
\label{fig_preservation_error}
\end{center}
\end{figure}

\linetitle{Irrelevance Preservation Error}
Irrelevance preservation error is to evaluate whether the attribute irrelevant details are kept after the editing, e.g., the skin color should not change when adding a mustache.
Since all faces are aligned, for each attribute, we define an irrelevant region that should not be altered when editing this attribute, as illustrated in Fig.~\ref{fig_mask_for_appearance_change} by examples (please refer to the supplementary material for the irrelevant region definition for all attributes).
Then the irrelevance preservation error is calculated as the L1 difference of the irrelevant region between the edited image and the original image, and the results are shown in Fig.~\ref{fig_preservation_error}.
As can be seen, the proposed PA-GAN achieves much lower preservation error than StarGAN, AttGAN, and STGAN.
Although SaGAN achieves slightly lower preservation error than PA-GAN, SaGAN has much lower attribute editing accuracy as seen from Fig.~\ref{fig_attribute_accuracy}, which means that SaGAN has a serious compromise between the attribute correctness and irrelevance preservation error.
As for the proposed PA-GAN, as seen from Fig.~\ref{fig_attribute_accuracy} and Fig.~\ref{fig_preservation_error}, PA-GAN achieves high attribute editing accuracy as well as considerably low preservation error, credited to the progressive attention mechanism.

\begin{figure}[!ht]
\begin{center}
\vspace{-5mm}
\includegraphics[width=1\linewidth]{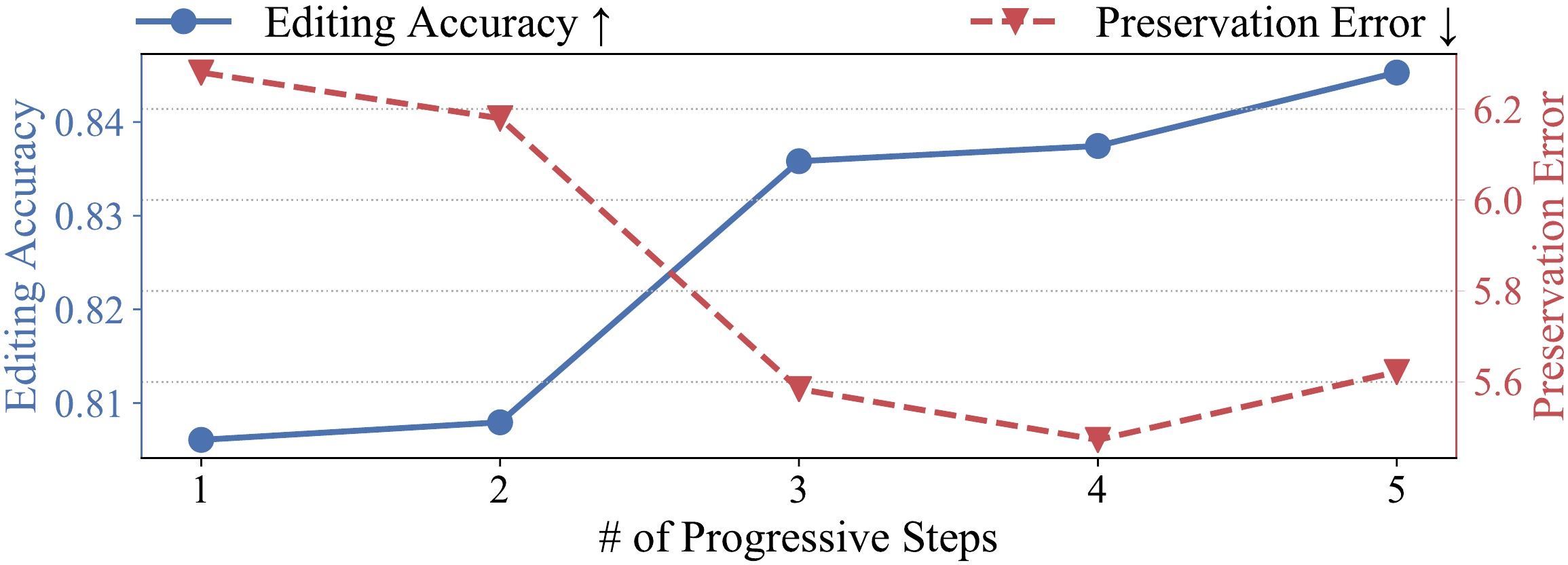}
\end{center}
\vspace{-4mm}
\caption{Effect of the number of progressive steps.}
\label{fig_accuracy_error_progressive_steps}
\end{figure}

\begin{table}[!ht]
\centering
\resizebox{\linewidth}{!}{%
\begin{tabular}{|c|c|c|c|c|c|}
\hline
                  & (a)    & (b)          & (c)         & (d)           & (e)             \\
                  & Full   & w/o Residual & Single Mask & w/o $l_{spa}$ & w/o $l_{ovl}$ \\ \hline\hline
Editing Acc.      & 83.7\% & 78.6\%       & 78.3\%      & 83.0\%        & 82.9\%          \\ \hline
Preservation Err. & 5.47   & 7.42         & 7.74        & 6.18          & 6.45            \\ \hline
\end{tabular}%
}
\caption{Ablation study by removing (b) residual strategy, (c) multiple masks, (d) mask sparsity loss, and (e) mask overlap loss respectively. (a) denotes our full PA-GAN.}\label{tab_effect}
\vspace{-3mm}
\end{table}

\subsection{Ablation Study}

\linetitle{Effect of Progressive Steps}
In Fig.~\ref{fig_accuracy_error_progressive_steps}, we investigate the effect of changing the number of the progressive steps, where we remove a progressive step by replacing the corresponding attentive editor by normal convolutional layers.
As seen from Fig.~\ref{fig_accuracy_error_progressive_steps}, the attribute accuracy increases along with adding the progressive steps while the preservation error decreases, which demonstrates the effectiveness of the progressive manner.

\linetitle{Effect of Residual Strategy}
We cancel the residual strategy in Eq.~(\ref{eq_m}) and (\ref{eq_dm}) to evaluate its effect; instead, we directly generate the mask without residual.
Compared to the full method (Table~\ref{tab_effect}(a)), canceling the residual strategy (Table~\ref{tab_effect}(b)) results in 5.1\% drops on the attribute accuracy and increases the irrelevance preservation error, demonstrating the necessity and effectiveness of the residual strategy.

\linetitle{Effect of Multiple Masks}
In our method, distinct masks for distinct attributes are learned for multiple attribute editing as designed in Sec.~\ref{sec_ext}.
In this part, we investigate the effect of this strategy by canceling it, and instead, we directly generate a whole mask for all target attributes.
Compared to the full method (Table~\ref{tab_effect}(a)), canceling this strategy (Table~\ref{tab_effect}(c)) results in 5.4\% drops on the attribute editing accuracy and increases the irrelevance preservation error.
Therefore, the multiple mask strategy can bring considerable improvement.

\linetitle{Effect of Mask Losses}
We respectively remove the sparsity loss $l_{spa}$ in Eq.~(\ref{eq_lspa}) and the overlap loss $l_{ovl}$ in Eq.~(\ref{eq_lovl}) to investigate their effect.
As seen from Table~\ref{tab_effect}(d) and (e), removing the sparsity loss $l_{spa}$ or the overlap $l_{ovl}$ drops the attribute editing accuracy and increases the irrelevance preservation error.
Therefore, both the sparsity loss and the overlap loss benefit the editing.

\section{Conclusion}

In this work, we propose a novel progressive attention GAN for facial attribute editing, which progressively edits the attributes in an attention manner from high to low feature level.
Credited to the progressive attention mechanism, the attribute editing is conducted in a coarse-to-fine manner, which is robust and precise.
Experiments on CelebA dataset demonstrate the superiority of our method compared to the state-of-the-art methods.
In future work, we will investigate whether and how the progressive attention mechanism can be extended to attribute editing for general objects such as color editing of a car.

{\small
\bibliographystyle{ieee_fullname}
\bibliography{egbib}

\begin{thebibliography}{10}\itemsep=-1pt

\bibitem{arjovsky2017wasserstein}
Martin Arjovsky, Soumith Chintala, and L{\'e}on Bottou.
\newblock Wasserstein gan.
\newblock In {\em International Conference on Machine Learning}, 2017.

\bibitem{chen2018facelet}
Ying-Cong Chen, Huaijia Lin, Michelle Shu, Ruiyu Li, Xin Tao, Xiaoyong Shen,
  Yangang Ye, and Jiaya Jia.
\newblock Facelet-bank for fast portrait manipulation.
\newblock In {\em IEEE Conference on Computer Vision and Pattern Recognition},
  2018.

\bibitem{chen2019semantic}
Ying-Cong Chen, Xiaohui Shen, Zhe Lin, Xin Lu, I Pao, Jiaya Jia, et~al.
\newblock Semantic component decomposition for face attribute manipulation.
\newblock In {\em IEEE Conference on Computer Vision and Pattern Recognition},
  2019.

\bibitem{chen2019homomorphic}
Ying-Cong Chen, Xiaogang Xu, Zhuotao Tian, and Jiaya Jia.
\newblock Homomorphic latent space interpolation for unpaired image-to-image
  translation.
\newblock In {\em IEEE Conference on Computer Vision and Pattern Recognition},
  2019.

\bibitem{choi2018stargan}
Yunjey Choi, Minje Choi, Munyoung Kim, Jung-Woo Ha, Sunghun Kim, and Jaegul
  Choo.
\newblock Stargan: Unified generative adversarial networks for multi-domain
  image-to-image translation.
\newblock In {\em IEEE Conference on Computer Vision and Pattern Recognition},
  2018.

\bibitem{goodfellow2014generative}
Ian Goodfellow, Jean Pouget-Abadie, Mehdi Mirza, Bing Xu, David Warde-Farley,
  Sherjil Ozair, Aaron Courville, and Yoshua Bengio.
\newblock Generative adversarial networks.
\newblock In {\em Advances in Neural Information Processing Systems}, 2014.

\bibitem{gulrajani2017improved}
Ishaan Gulrajani, Faruk Ahmed, Martin Arjovsky, Vincent Dumoulin, and Aaron
  Courville.
\newblock Improved training of wasserstein gans.
\newblock In {\em Advances in Neural Information Processing Systems}, 2017.

\bibitem{he2019attgan}
Zhenliang He, Wangmeng Zuo, Meina Kan, Shiguang Shan, and Xilin Chen.
\newblock Attgan: Facial attribute editing by only changing what you want.
\newblock {\em IEEE Transactions on Image Processing}, 2019.

\bibitem{karras2018progressive}
Tero Karras, Timo Aila, Samuli Laine, and Jaakko Lehtinen.
\newblock Progressive growing of gans for improved quality, stability, and
  variation.
\newblock In {\em International Conference on Learning Representations}, 2018.

\bibitem{karras2019style}
Tero Karras, Samuli Laine, and Timo Aila.
\newblock A style-based generator architecture for generative adversarial
  networks.
\newblock In {\em IEEE Conference on Computer Vision and Pattern Recognition},
  2019.

\bibitem{kim2019progressive}
Junyeong Kim, Minuk Ma, Kyungsu Kim, Sungjin Kim, and Chang~D Yoo.
\newblock Progressive attention memory network for movie story question
  answering.
\newblock In {\em IEEE Conference on Computer Vision and Pattern Recognition},
  2019.

\bibitem{kim2017unsupervised}
Taeksoo Kim, Byoungjip Kim, Moonsu Cha, and Jiwon Kim.
\newblock Unsupervised visual attribute transfer with reconfigurable generative
  adversarial networks.
\newblock {\em arXiv:1707.09798}, 2017.

\bibitem{kingma2015adam}
Diederik Kingma and Jimmy Ba.
\newblock Adam: A method for stochastic optimization.
\newblock In {\em International Conference on Learning Representations}, 2015.

\bibitem{kingma2013auto}
Diederik~P Kingma and Max Welling.
\newblock Auto-encoding variational bayes.
\newblock {\em arXiv:1312.6114}, 2013.

\bibitem{lample2017fader}
Guillaume Lample, Neil Zeghidour, Nicolas Usunier, Antoine Bordes, Ludovic
  Denoyer, and Marc'Aurelio Ranzato.
\newblock Fader networks: Manipulating images by sliding attributes.
\newblock In {\em Advances in Neural Information Processing Systems}, 2017.

\bibitem{larsen2016autoencoding}
Anders Boesen~Lindbo Larsen, S{\o}ren~Kaae S{\o}nderby, Hugo Larochelle, and
  Ole Winther.
\newblock Autoencoding beyond pixels using a learned similarity metric.
\newblock In {\em International Conference on Machine Learning}, 2016.

\bibitem{li2016convolutional}
Mu Li, Wangmeng Zuo, and David Zhang.
\newblock Convolutional network for attribute-driven and identity-preserving
  human face generation.
\newblock {\em arXiv:1608.06434}, 2016.

\bibitem{li2016deep}
Mu Li, Wangmeng Zuo, and David Zhang.
\newblock Deep identity-aware transfer of facial attributes.
\newblock {\em arXiv:1610.05586}, 2016.

\bibitem{liu2019stgan}
Ming Liu, Yukang Ding, Min Xia, Xiao Liu, Errui Ding, Wangmeng Zuo, and Shilei
  Wen.
\newblock Stgan: A unified selective transfer network for arbitrary image
  attribute editing.
\newblock In {\em IEEE Conference on Computer Vision and Pattern Recognition},
  2019.

\bibitem{liu2015deep}
Ziwei Liu, Ping Luo, Xiaogang Wang, and Xiaoou Tang.
\newblock Deep learning face attributes in the wild.
\newblock In {\em IEEE International Conference on Computer Vision}, 2015.

\bibitem{mao2017least}
Xudong Mao, Qing Li, Haoran Xie, Raymond~YK Lau, Zhen Wang, and Stephen
  Paul~Smolley.
\newblock Least squares generative adversarial networks.
\newblock In {\em IEEE International Conference on Computer Vision}, 2017.

\bibitem{mirza2014conditional}
Mehdi Mirza and Simon Osindero.
\newblock Conditional generative adversarial nets.
\newblock {\em arXiv:1411.1784}, 2014.

\bibitem{odena2016conditional}
Augustus Odena, Christopher Olah, and Jonathon Shlens.
\newblock Conditional image synthesis with auxiliary classifier gans.
\newblock In {\em Advances in Neural Information Processing Systems Workshops},
  2016.

\bibitem{perarnau2016invertible}
Guim Perarnau, Joost van~de Weijer, Bogdan Raducanu, and Jose~M {\'A}lvarez.
\newblock Invertible conditional gans for image editing.
\newblock In {\em Advances in Neural Information Processing Systems Workshops},
  2016.

\bibitem{petzka2018regularization}
Henning Petzka, Asja Fischer, and Denis Lukovnicov.
\newblock On the regularization of wasserstein gans.
\newblock {\em International Conference on Learning Representations}, 2018.

\bibitem{romero2019smit}
Andr{\'e}s Romero, Pablo Arbel{\'a}ez, Luc Van~Gool, and Radu Timofte.
\newblock Smit: Stochastic multi-label image-to-image translation.
\newblock In {\em IEEE International Conference on Computer Vision Workshops},
  2019.

\bibitem{ronneberger2015u}
Olaf Ronneberger, Philipp Fischer, and Thomas Brox.
\newblock U-net: Convolutional networks for biomedical image segmentation.
\newblock In {\em International Conference on Medical image computing and
  computer-assisted intervention}, 2015.

\bibitem{seo2016progressive}
Paul~Hongsuck Seo, Zhe Lin, Scott Cohen, Xiaohui Shen, and Bohyung Han.
\newblock Progressive attention networks for visual attribute prediction.
\newblock In {\em British Machine Vision Conference}, 2018.

\bibitem{shen2017learning}
Wei Shen and Rujie Liu.
\newblock Learning residual images for face attribute manipulation.
\newblock In {\em IEEE Conference on Computer Vision and Pattern Recognition},
  2017.

\bibitem{sungatullina2018image}
Diana Sungatullina, Egor Zakharov, Dmitry Ulyanov, and Victor Lempitsky.
\newblock Image manipulation with perceptual discriminators.
\newblock In {\em European Conference on Computer Vision}, 2018.

\bibitem{upchurch2017deep}
Paul Upchurch, Jacob Gardner, Geoff Pleiss, Robert Pless, Noah Snavely, Kavita
  Bala, and Kilian Weinberger.
\newblock Deep feature interpolation for image content changes.
\newblock In {\em IEEE Conference on Computer Vision and Pattern Recognition},
  2017.

\bibitem{wu2019relgan}
Po-Wei Wu, Yu-Jing Lin, Che-Han Chang, Edward~Y Chang, and Shih-Wei Liao.
\newblock Relgan: Multi-domain image-to-image translation via relative
  attributes.
\newblock In {\em IEEE International Conference on Computer Vision}, 2019.

\bibitem{xiao2018dna}
Taihong Xiao, Jiapeng Hong, and Jinwen Ma.
\newblock Dna-gan: Learning disentangled representations from multi-attribute
  images.
\newblock In {\em International Conference on Learning Representations
  Workshops}, 2018.

\bibitem{xiao2018elegant}
Taihong Xiao, Jiapeng Hong, and Jinwen Ma.
\newblock Elegant: Exchanging latent encodings with gan for transferring
  multiple face attributes.
\newblock In {\em European Conference on Computer Vision}, 2018.

\bibitem{yin2019instance}
Weidong Yin, Ziwei Liu, and Chen~Change Loy.
\newblock Instance-level facial attributes transfer with geometry-aware flow.
\newblock In {\em AAAI Conference on Artificial Intelligence}, 2019.

\bibitem{zhang2018generative}
Gang Zhang, Meina Kan, Shiguang Shan, and Xilin Chen.
\newblock Generative adversarial network with spatial attention for face
  attribute editing.
\newblock In {\em European Conference on Computer Vision}, 2018.

\bibitem{zhang2017stackgan}
Han Zhang, Tao Xu, Hongsheng Li, Shaoting Zhang, Xiaogang Wang, Xiaolei Huang,
  and Dimitris~N Metaxas.
\newblock Stackgan: Text to photo-realistic image synthesis with stacked
  generative adversarial networks.
\newblock In {\em IEEE International Conference on Computer Vision}, 2017.

\bibitem{zhang2018sparsely}
Jichao Zhang, Yezhi Shu, Songhua Xu, Gongze Cao, Fan Zhong, Meng Liu, and
  Xueying Qin.
\newblock Sparsely grouped multi-task generative adversarial networks for
  facial attribute manipulation.
\newblock In {\em ACM international conference on Multimedia}, 2018.

\bibitem{zhang2018progressive}
Xiaoning Zhang, Tiantian Wang, Jinqing Qi, Huchuan Lu, and Gang Wang.
\newblock Progressive attention guided recurrent network for salient object
  detection.
\newblock In {\em IEEE Conference on Computer Vision and Pattern Recognition},
  2018.

\bibitem{zhao2018modular}
Bo Zhao, Bo Chang, Zequn Jie, and Leonid Sigal.
\newblock Modular generative adversarial networks.
\newblock In {\em European Conference on Computer Vision}, 2018.

\bibitem{zhou2017genegan}
Shuchang Zhou, Taihong Xiao, Yi Yang, Dieqiao Feng, Qinyao He, and Weiran He.
\newblock Genegan: Learning object transfiguration and attribute subspace from
  unpaired data.
\newblock In {\em British Machine Vision Conference}, 2017.

\bibitem{zhu2017unpaired}
Jun-Yan Zhu, Taesung Park, Phillip Isola, and Alexei~A Efros.
\newblock Unpaired image-to-image translation using cycle-consistent
  adversarial networks.
\newblock In {\em IEEE International Conference on Computer Vision}, 2017.

\end{thebibliography}
}

\clearpage
\setcounter{section}{0}
\setcounter{page}{1}

\twocolumn[\centerline{\textbf{\Large Appendix}}\vspace{5mm}]

\section{Predefined Set of Disjoint Attribute Pairs}

The full definition of $S$ in Eq.~(16) of the manuscript is
$S = \{$
(Bald, Mouth Open),
(Bald, Mustache),
(Bald, Beard),
(Bangs, Mouth Open),
(Bangs, Mustache),
(Bangs, Beard),
(Black Hair, Mouth Open),
(Black Hair, Mustache),
(Black Hair, Beard),
(Blond Hair, Mouth Open),
(Blond Hair, Mustache),
(Blond Hair, Beard),
(Brown Hair, Mouth Open),
(Brown Hair, Mustache),
(Brown Hair, Beard),
(Bushy Eyebrows, Mustache),
(Bushy Eyebrows, Beard),
(Eyeglasses, Mustache),
(Eyeglasses, Beard)
$\}$.

\section{Predefined Attribute-Irrelevant Regions}

In Sec.~4.2 of the manuscript, we introduce the irrelevance preservation error, which is calculated as the L1 difference of a predefined attribute-irrelevant region between the editing result and the original image.
Fig.~\ref{fig_attribute_irrelevant_region} shows the predefined attribute-irrelevant regions for all attributes.

\begin{table}[!ht]
\renewcommand\arraystretch{0.65}
\centering
\resizebox{\linewidth}{!}{%
\begin{tabular}{|c|c|c|}
\hline
\multirow{2}{*}{$E$}                                                 & \multirow{2}{*}{$C$}                                                                                     & \multirow{2}{*}{$D$}                  \\
                                                                     &                                                                                                          & \\ \hline\hline
\multirow{2}{*}{Conv(64, 4, 2) $\rightarrow$ BN $\rightarrow$ ReLU}  & \multicolumn{2}{c|}{\multirow{2}{*}{Conv(64, 4, 2) $\rightarrow$ LN $\rightarrow$ LReLU}}             \\
                                                                     & \multicolumn{2}{c|}{}                                                         \\ \hline
\multirow{2}{*}{Conv(128, 4, 2) $\rightarrow$ BN $\rightarrow$ ReLU} & \multicolumn{2}{c|}{\multirow{2}{*}{Conv(128, 4, 2) $\rightarrow$ LN $\rightarrow$ LReLU}}            \\
                                                                     & \multicolumn{2}{c|}{}                                                         \\ \hline
\multirow{2}{*}{Conv(256, 4, 2) $\rightarrow$ BN $\rightarrow$ ReLU} & \multicolumn{2}{c|}{\multirow{2}{*}{Conv(256, 4, 2) $\rightarrow$ LN $\rightarrow$ LReLU}}            \\
                                                                     & \multicolumn{2}{c|}{}                                                         \\ \hline
\multirow{2}{*}{Conv(512, 4, 2) $\rightarrow$ BN $\rightarrow$ ReLU} & \multicolumn{2}{c|}{\multirow{2}{*}{Conv(512, 4, 2) $\rightarrow$ LN $\rightarrow$ LReLU}}            \\
                                                                     & \multicolumn{2}{c|}{}                                                         \\ \hline
\multirow{2}{*}{}                                                    & \multicolumn{2}{c|}{\multirow{2}{*}{Conv(1024, 4, 2) $\rightarrow$ LN $\rightarrow$ LReLU}}           \\
                                                                     & \multicolumn{2}{c|}{}                                                         \\ \hline
\multirow{2}{*}{}                                                    & \multirow{2}{*}{FC(1024) $\rightarrow$ LReLU}                                                            & \multirow{2}{*}{FC(1024) $\rightarrow$ LReLU} \\
                                                                     &                                                                                                          & \\ \hline
\multirow{2}{*}{}                                                    & \multirow{2}{*}{FC(13) $\rightarrow$ Sigmoid}                                                                                   & \multirow{2}{*}{FC(1)}     \\
                                                                     &                                                                                                          & \\ \hline
\end{tabular}%
}
\caption{
Network architectures of the encoder $E$, the attribute classifier $C$, and the discriminator $D$.
$C$ and $D$ share most layers except for the last two layers.
}\label{tab_net_edc}
\end{table}

\begin{table}[!ht]
\renewcommand\arraystretch{0.65}
\centering
\vspace{-5mm}
\resizebox{\linewidth}{!}{%
\begin{tabular}{|c|c|}
\hline
\multirow{2}{*}{$G_{e_k}(k\neq0)$}                             & \multirow{2}{*}{$G_{e_0}$}                    \\
                                                               &                                               \\ \hline\hline
\multirow{2}{*}{DeConv(64$\times$2$^{k-1}$, 3, 1) $\rightarrow$ BN $\rightarrow$ ReLU} & \multirow{2}{*}{DeConv(32, 3, 1) $\rightarrow$ BN $\rightarrow$ ReLU} \\
                                                               &                                               \\ \hline
\multirow{2}{*}{DeConv(64$\times$2$^{k-1}$, 3, 2) $\rightarrow$ BN $\rightarrow$ ReLU} & \multirow{2}{*}{DeConv(3, 3, 2) $\rightarrow$ Tanh}  \\
                                                               &                                               \\ \hline
\end{tabular}%
}
\caption{Network architecture of $G_{e_k}$.}\label{tab_net_gek}
\end{table}

\begin{table}[!ht]
\renewcommand\arraystretch{0.65}
\centering
\vspace{-5mm}
\resizebox{\linewidth}{!}{%
\begin{tabular}{|c|}
\hline
\multirow{2}{*}{Attribute Predictor}                                                  \\
                                                                                      \\ \hline\hline
\multirow{2}{*}{Conv(16, 3, 1) $\rightarrow$ BN $\rightarrow$ ReLU $\rightarrow$ Conv(16, 3, 1) $\rightarrow$ BN $\rightarrow$ ReLU $\rightarrow$ Pool(2, 2)} \\
                                                                                      \\ \hline
\multirow{2}{*}{Conv(32, 3, 1) $\rightarrow$ BN $\rightarrow$ ReLU $\rightarrow$ Conv(32, 3, 1) $\rightarrow$ BN $\rightarrow$ ReLU $\rightarrow$ Pool(2, 2)} \\
                                                                                      \\ \hline
\multirow{2}{*}{Conv(64, 3, 1) $\rightarrow$ BN $\rightarrow$ ReLU $\rightarrow$ Conv(64, 3, 1) $\rightarrow$ BN $\rightarrow$ ReLU $\rightarrow$ Pool(2, 2)} \\
                                                                                      \\ \hline
\multirow{2}{*}{Conv(128, 3, 1) $\rightarrow$ BN $\rightarrow$ ReLU $\rightarrow$ Conv(128, 3, 1) $\rightarrow$ BN $\rightarrow$ ReLU $\rightarrow$ Pool(2, 2)} \\
                                                                                      \\ \hline
\multirow{2}{*}{FC(512) $\rightarrow$ ReLU}                                                       \\
                                                                                      \\ \hline
\multirow{2}{*}{FC(40) $\rightarrow$ Sigmoid}                                                    \\
                                                                                      \\ \hline
\end{tabular}%
}
\caption{Network architecture of the attribute predictor for the evaluation of the attribute editing accuracy.}\label{tab_attribute_predictor}
\end{table}

\begin{table*}[!b]
\renewcommand\arraystretch{0.65}
\centering
\vspace{-5mm}
\resizebox{\linewidth}{!}{%
\begin{tabular}{|c|c|l|c|}
\hline
\multicolumn{4}{|c|}{\multirow{2}{*}{$G_{m_k}$}}                                                                                                                                                                                                     \\
\multicolumn{4}{|c|}{}                                                                                                                                                                                                                         \\ \hline\hline
\multirow{6}{*}{Conv(64$\times$2$^{k-1}$, 1, 1) $\rightarrow$ BN $\rightarrow$ ReLU} & \multirow{6}{*}{Conv(64$\times$2$^{k-1}$, 3, 1) $\rightarrow$ BN $\rightarrow$ ReLU} & \multicolumn{1}{c|}{\multirow{3}{*}{Conv(64$\times$2$^{k-1}$, 3, 1) $\rightarrow$ BN $\rightarrow$ ReLU}} & \multirow{2}{*}{Conv(64$\times$2$^{k-1}$, 3, 1) $\rightarrow$ BN $\rightarrow$ ReLU} \\
                                                                                     &                                                                                      & \multicolumn{1}{c|}{}                                                                                     & \\ \cline{4-4}
                                                                                     &                                                                                      & \multicolumn{1}{c|}{}                                                                                     & \multirow{2}{*}{Conv(64$\times$2$^{k-1}$, 3, 1) $\rightarrow$ BN $\rightarrow$ ReLU} \\ \cline{3-3}
                                                                                     &                                                                                      & \multirow{3}{*}{Conv(64$\times$2$^{k-1}$, 3, 1) $\rightarrow$ BN $\rightarrow$ ReLU}                      & \\ \cline{4-4}
                                                                                     &                                                                                      &                                                                                                           & \multirow{2}{*}{Conv(64$\times$2$^{k-1}$, 3, 1) $\rightarrow$ BN $\rightarrow$ ReLU} \\
                                                                                     &                                                                                      &                                                                                                           & \\ \hline
\multicolumn{4}{|c|}{\multirow{2}{*}{Concatenation}}                                                                                                                                                                                                  \\
\multicolumn{4}{|c|}{}                                                                                                                                                                                                                         \\ \hline
\multicolumn{4}{|c|}{\multirow{2}{*}{Conv(64$\times$2$^k$, 4, 2) $\rightarrow$ BN $\rightarrow$ ReLU}}                                                                                                                                                                   \\
\multicolumn{4}{|c|}{}                                                                                                                                                                                                                         \\ \hline
\multicolumn{4}{|c|}{\multirow{2}{*}{DeConv(13, 4, 2)}}                                                                                                                                                                                        \\
\multicolumn{4}{|c|}{}                                                                                                                                                                                                                         \\ \hline
\end{tabular}%
}
\caption{Network architecture of $G_{m_k}$.}\label{tab_net_gmk}
\end{table*}

\section{Network Architectures}

Table~\ref{tab_net_edc} shows the architectures of the encoder $E$, the attribute classifier $C$, and the discriminator $D$.
Table~\ref{tab_net_gek} shows the architecture of $G_{e_k}$, and Table~\ref{tab_net_gmk} shows the architecture of $G_{m_k}$.
Table~\ref{tab_attribute_predictor} shows the architecture of the attribute predictor for the evaluation of the attribute editing accuracy (see Sec.~4.2 of the manuscript).
In these tables, Conv(c,~k,~s) and DeConv(c,~k,~s) respectively denote the convolutional layer and the transposed convolutional layer with c as output channels, k as kernel size, and s as stride.
FC(d) denotes the fully connected layer with d as dimension.
Pool(k,~s) denotes the max-pooling layer with k as kernel size and s as stride.
%
BN denotes the batch normalization and LN denotes the layer normalization. LReLU denotes the leaky ReLU.

\section{Additional Visual Results}

\begin{figure*}[!ht]
\begin{center}
   \includegraphics[width=1\linewidth]{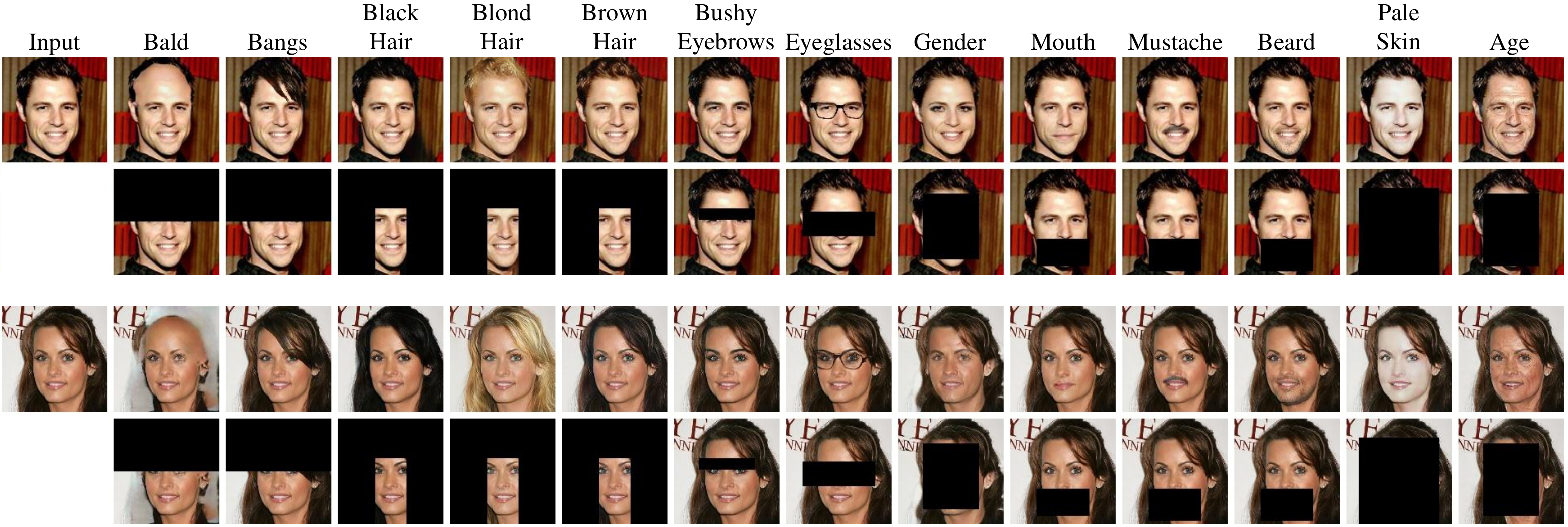}
\end{center}
   \vspace{-4mm}
   \caption{
   Predefined attribute-irrelevant regions.
   For each attribute, the non-black region is defined as the attribute-irrelevant region, which should not be changed when editing the corresponding attribute.
   The irrelevance preservation error is calculated as the L1 difference of the non-black region between the editing result and the input image.
   }
   \label{fig_attribute_irrelevant_region}

\begin{center}
   \includegraphics[width=1\linewidth]{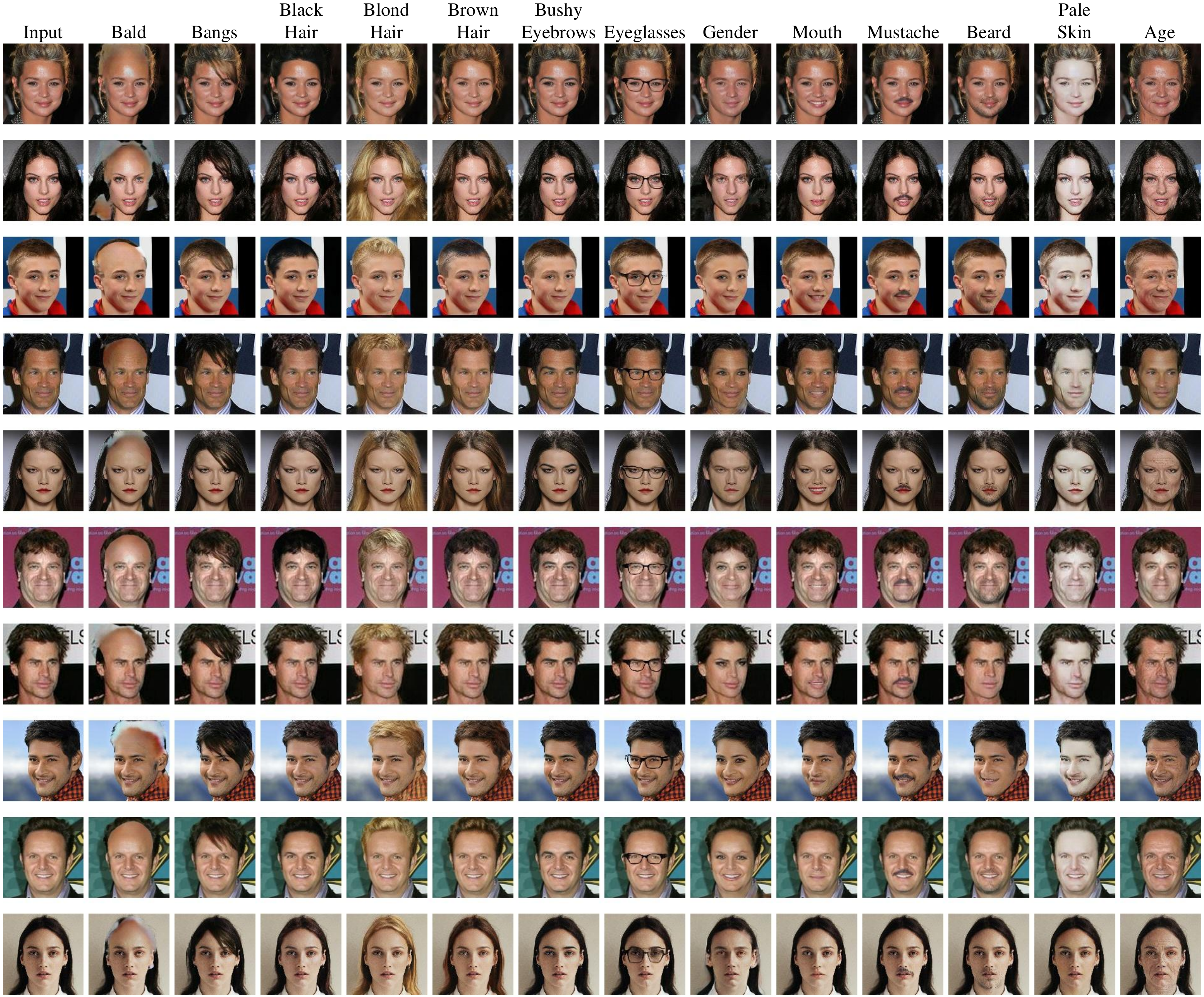}
\end{center}
   \vspace{-4mm}
   \caption{
   Attribute editing on 128$\times$128 images.
   We \textit{invert} the state of each specified attribute, e.g, to edit female to male, male to female, to remove the existing beard, and to add a beard if it does not exist.
   }
   \label{fig_128}
\end{figure*}

\begin{landscape}
\begin{figure}[p]
\begin{center}
   \includegraphics[width=1\linewidth]{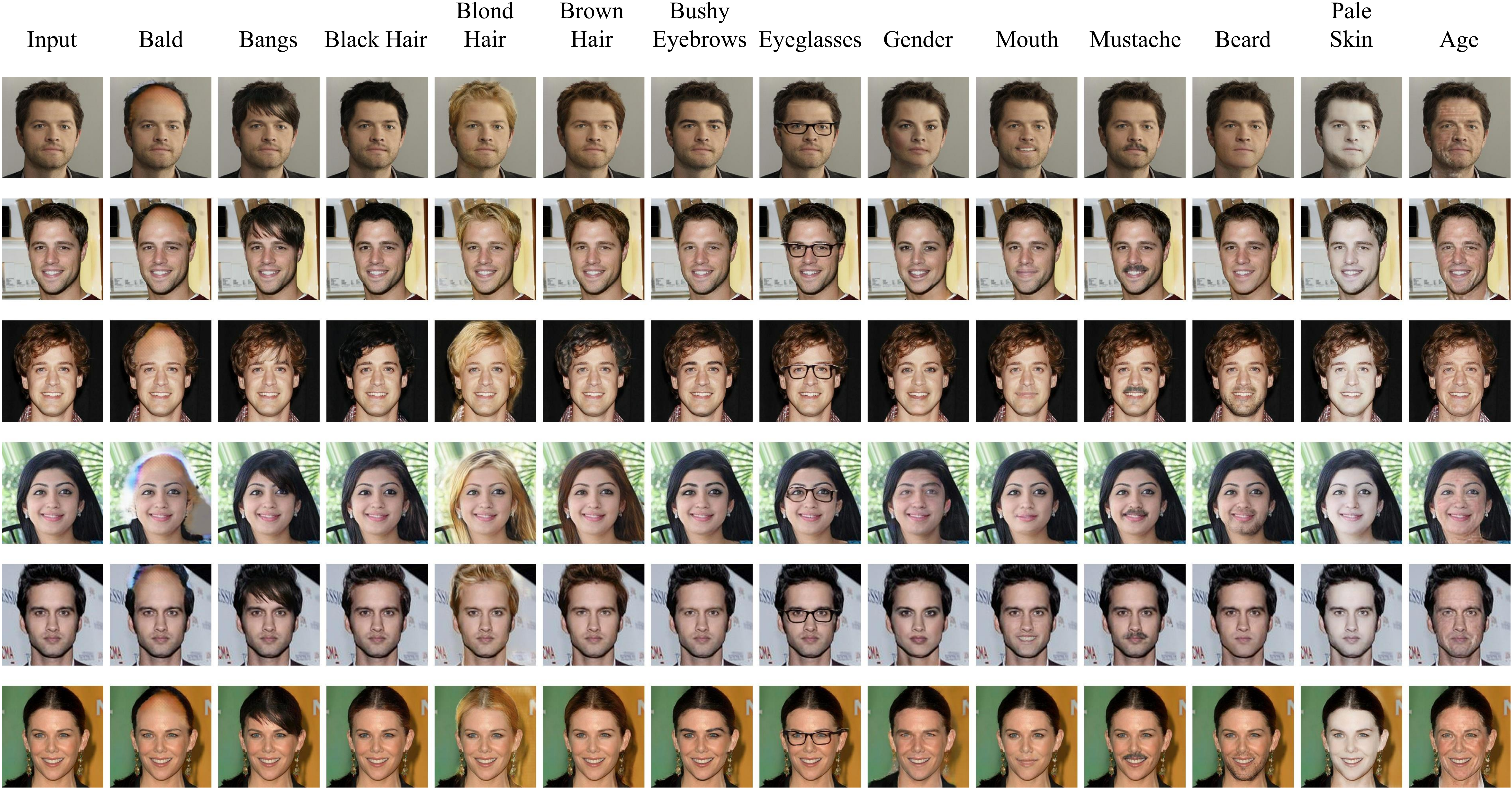}
\end{center}
   \caption{
   Attribute editing on 256$\times$256 images.
   We \textit{invert} the state of each specified attribute, e.g, to edit female to male, male to female, to remove the existing beard, and to add a beard if it does not exist.
   Please zoom in for a better view.
   }
   \label{fig_256}
\end{figure}
\end{landscape}

\begin{landscape}
\begin{figure}[p]
\begin{center}
   \includegraphics[width=\linewidth]{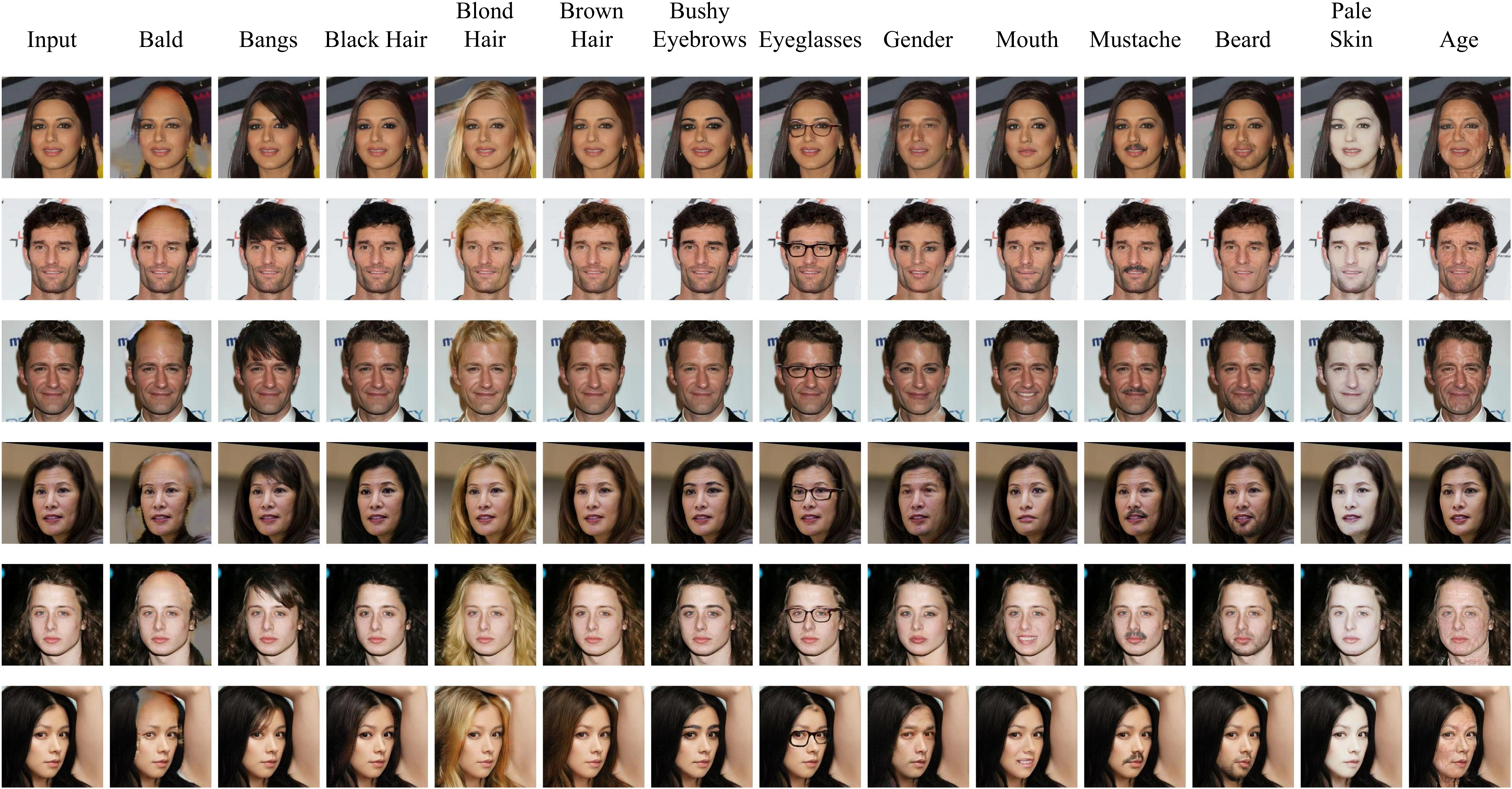}
\end{center}
   \caption{
   Attribute editing on 256$\times$256 images.
   We \textit{invert} the state of each specified attribute, e.g, to edit female to male, male to female, to remove the existing beard, and to add a beard if it does not exist.
   Please zoom in for a better view.
   }
   \label{fig_256_2}
\end{figure}
\end{landscape}

\begin{landscape}
\begin{figure}[p]
\begin{center}
   \includegraphics[width=\linewidth]{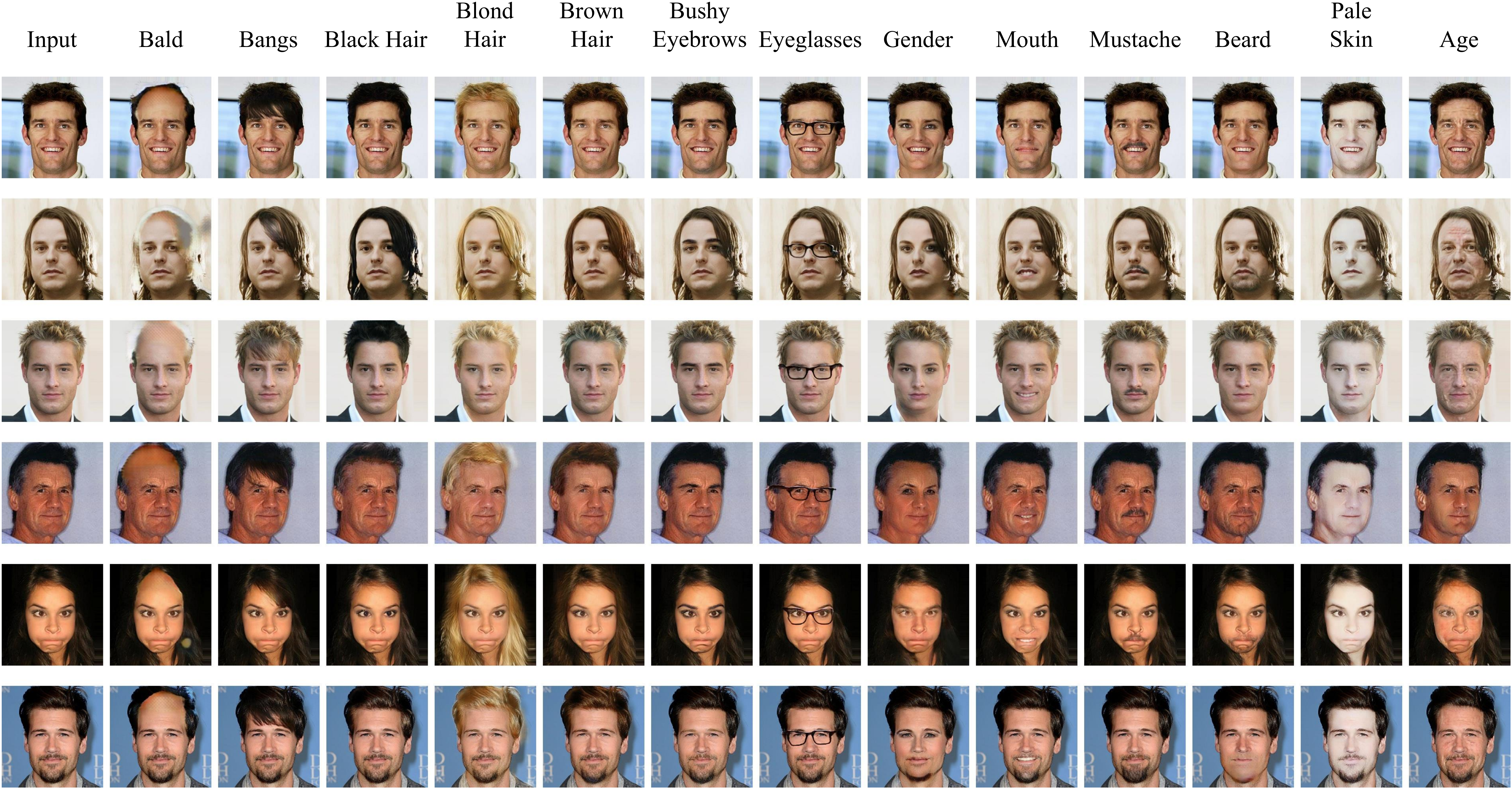}
\end{center}
   \caption{
   Attribute editing on 256$\times$256 images.
   We \textit{invert} the state of each specified attribute, e.g, to edit female to male, male to female, to remove the existing beard, and to add a beard if it does not exist.
   Please zoom in for a better view.
   }
   \label{fig_256_3}
\end{figure}
\end{landscape}

Additional visual results of 128$\times$128 images are shown in Fig.~\ref{fig_128}, and the results of 256$\times$256 images are shown in Fig.~\ref{fig_256}, Fig~\ref{fig_256_2}, and Fig~\ref{fig_256_3}.
As can be seen, overall, our method can generate correct attributes with high fidelity and well preserve the irrelevance.
Even for the most challenging ``Bald'' attribute which is hard for all methods as analyzed in Sec.~4.2 of the manuscript, although artifacts appear in some cases, our method can still produce some satisfactory ``Bald'' results.

\clearpage
\pagenumbering{gobble}
\newgeometry{top=0.5cm,bottom=0.5cm}

\end{document}